\documentclass[12pt]{article}
\usepackage{arxiv}
\usepackage{graphicx,epstopdf}
\usepackage{amsmath}
\usepackage{times}
\usepackage{graphicx}
\usepackage{color}
\usepackage{multirow}
\usepackage{mathtools}
\usepackage{amssymb}
\usepackage{booktabs}       
\usepackage{amsfonts}       
\usepackage{nicefrac}       
\usepackage[authoryear]{natbib}
\usepackage{url}

\title{The two-dimensional Gabor function adapted to natural image statistics: A model of simple-cell receptive fields and sparse structure in images}
\date{}

\author{P.~N.~Loxley\\
School of Science and Technology,\\ 
University of New England,\\ Armidale, NSW, Australia.}

\renewcommand{\headeright}{A Postprint}
\renewcommand{\undertitle}{A Postprint}

\begin{document}
\maketitle

\begin{abstract}
The two-dimensional Gabor function is adapted to natural image statistics, leading to a tractable probabilistic generative model that can be used to model simple-cell receptive-field profiles, or generate basis functions for sparse coding applications. Learning is found to be most pronounced in three Gabor-function parameters representing the size and spatial frequency of the two-dimensional Gabor function, and characterized by a non-uniform probability distribution with heavy tails. All three parameters are found to be strongly correlated: resulting in a basis of multiscale Gabor functions with similar aspect ratios, and size-dependent spatial frequencies. A key finding is that the distribution of receptive-field sizes is scale-invariant over a wide range of values, so there is no characteristic receptive-field size selected by natural image statistics. The Gabor-function aspect ratio is found to be approximately conserved by the learning rules and is therefore not well-determined by natural image statistics. This allows for three distinct solutions: a basis of Gabor functions with sharp orientation resolution at the expense of spatial-frequency resolution; a basis of Gabor functions with sharp spatial-frequency resolution at the expense of orientation resolution; or a basis with unit aspect ratio. Arbitrary mixtures of all three cases are also possible. Two parameters controlling the shape of the marginal distributions in a probabilistic generative model fully account for all three solutions. The best-performing probabilistic generative model for sparse coding applications is found to be a Gaussian copula with Pareto marginal probability density functions. 
\end{abstract}

\section{Introduction}\label{intro}

We know that simple cells in the primary visual cortex have spatially-localized receptive fields and are sensitive to visual stimulus features such as orientation, spatial frequency, and location in the receptive field \citep{hubel1,hubel2,daugman85,jones}. Simple cells also help map visual stimulus features from the visual field to a retinotopic position on the surface of the primary visual cortex \citep{durbin,swindale}. Simple-cell receptive-field profiles have been extensively studied and modeled using the Gabor function \citep{marcelja,daugman85,jones}, and various models explaining receptive field development from visual experience have been proposed. 

Early neural network models described the self-organization of orientation selective cells through Hebbian learning with localized, oriented, input patterns \citep{vondermalsburg}, and uncorrelated random input \citep{linsker}. More recent models made use of realistic inputs given by natural sensory data. Implicit in these models is the assumption that the early visual system has evolved to process natural image statistics. Natural sensory data contains redundancy, leading Attneave and Barlow to hypothesize that sensory systems re-code sensory data in a way that reduces redundancy; known as the \emph{efficient coding hypothesis}. Models capable of efficient coding of natural images (i.e., natural sensory data at the level of photoreceptors in the retina) have since been shown to develop realistic cortical receptive-field profiles \citep{olshausen96,olshausen97,bell97,hyvarinen00,hyvarinen01,sommer,olshausen13b}. Redundancy in natural images takes the form of statistical dependencies between pixels, and efficient coding models attempt to find a transformation that reduces these statistical dependencies. More specifically, models that learn realistic simple-cell receptive-field profiles reduce high-order statistical dependencies: second-order dependencies given by pairwise pixel correlations are assumed to have been removed at an earlier stage by some form of ``whitening" transform. Earlier work had already demonstrated that pixel dependencies in images could be reduced by applying the two-dimensional (2D) Gabor function as a wavelet transform, yielding an image code with a sparse distribution and a low marginal entropy \citep{daugman85,field87,daugman88,daugman89,lee}. Similarly, learning a sparse code for natural images results in a set of Gabor-like basis functions that can efficiently reconstruct natural images, so that a relatively small number of basis functions are responsible for most of the reconstruction of any given image.

Sparse coding and redundancy reduction serve purposes other than data compression \citep{foldiak,field94,eichhorn,hyvarinenbook}. One advantage of a sparse code is representational capacity \citep{foldiak}. Natural images are comprised of regions of related pixels, corresponding to objects, for example; while pixels associated with different objects have little statistical dependence on one another \citep{ruderman}. A sparse code attempts to represent such ``objects" as statistically independent components \citep{bell97}, allowing different images to be coded as different combinations of these independent components. A sparse code also implies a good statistical model of the data, allowing for tasks such as density estimation; denoising; or the generation of synthetic data resembling natural images \citep{eichhorn,hyvarinenbook}. Another advantage of a sparse code is energy efficiency. Encoding images using a sparse code can be done at low power as only small numbers of neurons are required to be active at any one time. These advantages all rely on the ability of simple cells to reduce redundancy in natural images. However, there is considerable debate as to how much redundancy actually exists in high-order pixel statistics \citep{li,petrov,eichhorn,hosseini}, and whether simple-cell receptive fields are actually capable of reducing this redundancy \citep{bethge,eichhorn}. Whatever the potential advantages are, the precise reason why simple cells would employ a sparse code is not yet clear. 

The aim of this work is to present an alternative approach to understanding simple-cell receptive fields, which is to quantify what is actually learned by a sparse coding model. One way to achieve this is to parameterize the receptive-field profiles, then quantify the statistical properties of these parameters after adapting to natural image statistics. Presumably, properties of the receptive-field profiles will reflect important properties of natural images. Perhaps the most robust and striking property of natural images is scale invariance. This has been confirmed in the second-order statistics of ensembles of images \citep{ruderman2,ruderman,mumford}, and in the high-order statistics of histograms of wavelet coefficients \citep{mumford}. Other properties assumed to be essential to natural images include the locality of objects, the linear superposition of clutter, and the existence of blank spaces in images \citep{mumford}. \cite{ruderman} showed that two sufficient properties for scale invariance in natural images are: (1) natural images are primarily composed of statistically-independent ``objects" which occlude one another, and  (2) image regions corresponding to these objects have a power-law distribution of sizes. It will be shown here that both of these properties are also present in natural images generated from learned receptive-field profiles in a sparse coding model. 

To quantify what is learned in a sparse coding model, I adapt the 2D Gabor function to natural image statistics by learning a joint probability distribution for the Gabor-function parameters. The structure of simple-cell receptive-field profiles can then be quantified through a detailed analysis of the statistical properties of the Gabor-function parameters. I also propose a tractable generative model by approximating the learned joint probability distribution. Non-parameterized approaches to sparse coding suggest that receptive-field profiles have the approximate form of 2D Gabor functions in the case of a modestly over-complete basis \citep{olshausen13b}. A highly over-complete basis allows for more diverse receptive-field profiles \citep{sommer,olshausen13b}. Therefore, in this work I consider the modestly over-complete case, and statistical properties are quantified using many data samples. 

The structure of this Article is as follows: In Sec.~2, the 2D Gabor function is parameterized in a way that allows it to adapt to natural image statistics. In Sec.~3, the results of adapting the 2D Gabor function to natural image statistics are presented. In Sec.~4, a probabilistic generative model for the Gabor-function parameters is derived and tested.  In Sec.~5, a brief summary and discussion are given. Learning rules for the Gabor-function parameters are presented in the Appendix.

\section{Model}\label{model}

The model presented here allows the 2D Gabor function to be adapted to natural image statistics by adjusting the Gabor-function parameters in an effort to find an optimal sparse code for natural images. I start with an image model given by a linear sum of basis functions $g({\bf{r}},{\bf{r}}^{\prime})$, and Gaussian noise $N({\bf{r}})$,
\begin{equation}
I({\bf{r}})=\sum_{{\bf{r}}^{\prime}}g({\bf{r}},{\bf{r}}^{\prime}) a({\bf{r}}^{\prime})+N({\bf{r}}),\label{gen}
\end{equation}
where ${\bf{r}}=(x,y)$ labels the discrete coordinates of image pixels $I({\bf{r}})$, and ${\bf{r}}^{\prime}=(x^\prime,y^\prime)$ labels the discrete coordinates of simple-cell activities $a({\bf{r}}^{\prime})$ forming the image representation. Each simple cell is labelled by ${\bf{r}}^{\prime}$, and corresponds to a unique basis function $g({\bf{r}},{\bf{r}}^{\prime})$. The sum is over the total number of basis functions (simple-cell labels). 

In order to investigate simple-cell receptive-field profiles, I choose a parameterized form for the basis functions that is motivated by the work of \cite{daugman85}, and \cite{jones}. I start with the real part of the 2D Gabor function in complex-exponential form:
\begin{equation}
G(x,y)=A\exp{\left[-\frac{1}{2}\left(\frac{\tilde{x}^2}{\sigma_{x}^2}+\frac{\tilde{y}^2}{\sigma_{y}^2}\right)\right]}\cos{\left(k\tilde{y}+\varphi\right)},
\label{gab1}
\end{equation}
with
\begin{equation}
(\tilde{x},\tilde{y})=
\left(\begin{array}{cc}
\cos{\phi}&-\sin{\phi}\\
\sin{\phi}&\cos{\phi}\\
\end{array}
\right)
\left(\begin{array}{c}
x-x_{0}\\
y-y_{0}\\
\end{array}
\right).
\label{gab2}
\end{equation}\\
This function has eight adjustable parameters and describes a 2D Gaussian modulated by a sinusoid. The Gaussian principal axes $\tilde{x}$ and $\tilde{y}$ are arbitrarily rotated and translated in the plane according to Eq.(\ref{gab2}). In this parameterization, the wave-vector of the sinusoidal term is always aligned along one of the principal axes; resulting in one less adjustable parameter than that presented in \cite{daugman85} and \cite{jones}. Specifically, I have chosen the wave-vector to be aligned along the $\tilde{y}$ principal axis. The 2D Gaussian envelope is described by its center location $(x_{0},y_{0})$, the envelope widths $\sigma_{x}$ and $\sigma_{y}$, and the orientation $\phi$ of its principal axes. The sinusoid has wavelength $\lambda=2\pi/|k|$ (or spatial frequency $f=|k|/2\pi$), phase $\varphi$, and a wave-vector along the $\tilde{y}$ principal axis. The parameter $A$ is a scale factor.

From the perspective of neuroscience, each Gabor parameter tells us something about the receptive-field profile of a cortical simple cell. The size of a receptive field located at $(x_0,y_0)$ is determined by $\sigma_{x}$ and $\sigma_{y}$. The parameter $\phi$ is the orientation of a visual stimulus given by a bar or an edge placed in a receptive field that elicits a strong response from the cell. The parameter $k$ indicates the preference of a cell for visual stimuli of particular spatial frequencies. Due to the parameterization of the wave-vector with respect to the principal axes, a receptive-field profile with an aspect ratio less than one, $\sigma_{y}/\sigma_{x}<1$, favours orientation resolution at the expense of spatial-frequency resolution; while a receptive-field profile with an aspect ratio greater than one, $\sigma_{y}/\sigma_{x}>1$, favours spatial-frequency resolution at the expense of orientation resolution \citep{daugman85}. This point is further investigated in the next section.

Each basis function is now chosen to have the form of a 2D Gabor function:\\
\begin{equation}
g({\bf{r}},{\bf{r}}^{\prime})=A\exp{\left[-\frac{1}{2}\left(\frac{\tilde{x}^2}{\sigma_{x}({\bf{r}}^{\prime})^2}+\frac{\tilde{y}^2}{\sigma_{y}({\bf{r}}^{\prime})^2}\right)\right]}\cos{\left[k({\bf{r}}^{\prime})\tilde{y}+\varphi({\bf{r}}^{\prime})\right]},\label{dic1}
\end{equation}
with
\begin{equation}
(\tilde{x},\tilde{y})=\left(\begin{array}{cc}
\cos{\phi({\bf{r}}^{\prime})}&-\sin{\phi({\bf{r}}^{\prime})}\\
\sin{\phi({\bf{r}}^{\prime})}&\cos{\phi({\bf{r}}^{\prime})}\\
\end{array}
\right)
\left(\begin{array}{c}
x-x_0({\bf{r}}^{\prime})\\
y-y_0({\bf{r}}^{\prime})\\
\end{array}
\right),\label{dic2}
\end{equation}\\
so that seven of the Gabor-function parameters in Eqs.~(\ref{dic1}) and (\ref{dic2}) now depend on ${\bf{r}}^{\prime}$. This means that a basis function with label ${\bf{r}}^{\prime}$ is described by the unique set of parameter values: ${\boldsymbol{\theta}}({\bf{r}}^{\prime})=(\phi({\bf{r}}^{\prime}),\varphi({\bf{r}}^{\prime}),\sigma_{x}({\bf{r}}^{\prime}),\sigma_{y}({\bf{r}}^{\prime}),k({\bf{r}}^{\prime}),x_0({\bf{r}}^{\prime}),y_0({\bf{r}}^{\prime}))$. The scale factor $A$ is chosen to be common to all basis functions (i.e., it is not a function of ${\bf{r}}^{\prime}$). 

It is important to note that I have chosen a set of basis functions to be 2D Gabor functions. However, a simple cell's receptive field is determined from the response of a Gabor \emph{filter} \citep{daugman85,jones}. It is demonstrated in \cite{olshausen96} that simple-cell receptive-field profiles are very similar in form to learned basis functions. Further, in \cite{hyvarinenbook}, it
 is shown that simple-cell receptive-field profiles actually correspond to low-pass-filtered basis functions with the same orientation, location, and spatial frequency tuning properties. In the following work, I therefore assume the parameters describing receptive-field profiles are equivalent to the parameters describing basis functions.

A learning rule for adjusting the Gabor-function parameters must now be derived. This is carried out in the Appendix and makes use of the EM algorithm \citep{dempster}. In the E step, the value of $a({\bf{r}})$ is inferred given ${\boldsymbol{\theta}}({\bf{r}}^{\prime})$ and $I({\bf{r}})$ using its MAP estimate:
\begin{equation}
\hat{a}({\bf{r}})=\underset{a({\bf{r}})}{\operatorname{arg\ min}}\sum_{\bf{r}}\left\{\frac{1}{2}\left[I({\bf{r}})-\sum_{{\bf{r}}^{\prime}}g({\bf{r}},{\bf{r}}^{\prime})a({\bf{r}}^{\prime})\right]^{2}+\nu S(a({\bf{r}}))\right\}.\label{objective2}\\
\end{equation}
In the M step, each Gabor-function parameter $\theta_{i}({\bf{r}})$ is then adjusted according to the update rule: 
\begin{equation}
\Delta\theta_{i}({\bf{r}})=\eta_{i} \sum_{{\bf{r}}^{\prime}} \frac{\partial g^{T}({\bf{r}},{\bf{r}^{\prime}})}{\partial\theta_{i}({\bf{r}})} \left\langle \hat{a}({\bf{r}})\hat{r}({\bf{r}}^{\prime})\right\rangle,\label{maplearn2}
\end{equation}
where $\hat{a}({\bf{r}})$ comes from the E step, $\hat{r}({\bf{r}}^{\prime})$ is the residual error, and the expectation is an average over a batch of images. There are some additional update steps required due to the MAP approximation (see Appendix for details). Results from applying these learning rules are discussed in the next section.

\section{Results}\label{results}

The Gabor-function parameters in Eqs.~(\ref{dic1}) and (\ref{dic2}) were adapted to natural image statistics by applying the learning rules derived in the Appendix. Ten natural images were chosen from the \emph{McGill Calibrated Colour Image Database} \citep{olmos}, and include images from the categories for flowers, foliage, landscapes, textures, and shadows. The images were firstly converted to grayscale, and then underwent whitening and dimensionality reduction according to the method described in \cite{olshausen97}. Batches of 100 image-patches of $16\times 16$ pixels were extracted randomly from these images and used to evaluate the expectation in Eq.~(\ref{maplearn2}). A total of 20000 parameter updates (image batches) were used.

The first objective is to show that application of the learning rules to natural images leads to a set of basis functions that are oriented, spatially localized, and bandpass (localized in the spatial-frequency domain); as originally shown in \cite{olshausen96}. To do this, I initialized the Gabor-function parameters to: $A=0.2$, $\phi=0$, $\varphi=0$, $\sigma_x=0.5$, $\sigma_y=0.5$, and $\lambda=0.23$. The parameters $x_0$ and $y_0$ where chosen to uniformly tile a square the size of an image patch, with $x_0\in(-0.5,0.5)$ and $y_0\in(-0.5,0.5)$. The spatial parameters are in dimensionless units, and should be multiplied by 16 (the side-length of an image patch) to convert to pixel units. This choice of parameter initialization was essentially arbitrary and uninformed, except for the initial values of $A$ and $\lambda$; which exhibited sensitivities and were found by trial and error. The choice of learning rate for $x_0$ and $y_0$ was $\eta=0.1$, while all other parameters used $\eta=1$. It was found that direct iteration of the learning rules did not give stable results. Although the EM algorithm is guaranteed to converge to a local maximum of the likelihood function, adding MAP heuristics to the E step results in additional updates (see Appendix) that potentially break this guarantee. After some experimentation a schedule of alternating between parameter updates, so that only one of the parameter groupings $(\sigma_x,x_0)$, $(\sigma_y,y_0)$, or $(\phi,\varphi,k)$ was updated in an arbitrary M step, was found to lead to a stable solution. 

\begin{figure}
\center
\includegraphics[width=250pt,bb=68 68 585 535, clip=true]{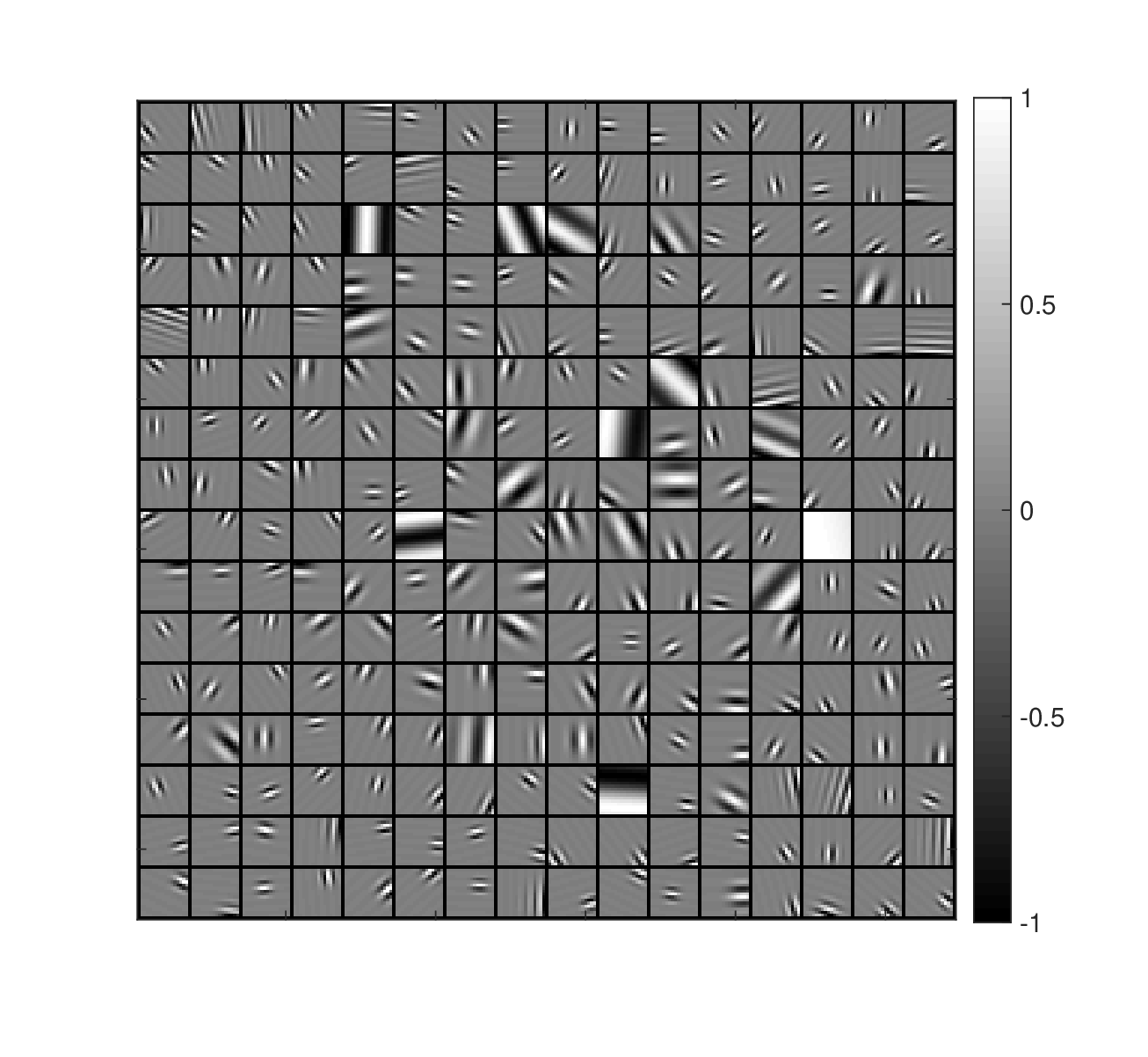}
\caption{An over-complete basis of 256 Gabor functions learned from natural images.}
\label{natural_basis}
\end{figure}

The result of applying the learning rules is shown in Fig.~\ref{natural_basis} for an initial basis of 256 Gabor functions (the number of Gabor functions was chosen to match the number of pixels in a $16\times 16$ image patch). It is clear from this figure that the learned basis functions are spatially localized (excepting the low-frequency cases). They are also bandpass and oriented. As a result of the dimensionality reduction performed during the image pre-processing stage this basis is over-complete (i.e., it spans the image space but the basis functions are linearly dependent). A singular-value decomposition shows there are 207 singular values greater than or equal to 0.01, while the remaining 49 are less than 0.01. Therefore, within a rank approximation, this basis is 1.2-times over-complete.

\begin{figure}
\center
\includegraphics[width=250pt,bb=13 45 376 763,clip=true]{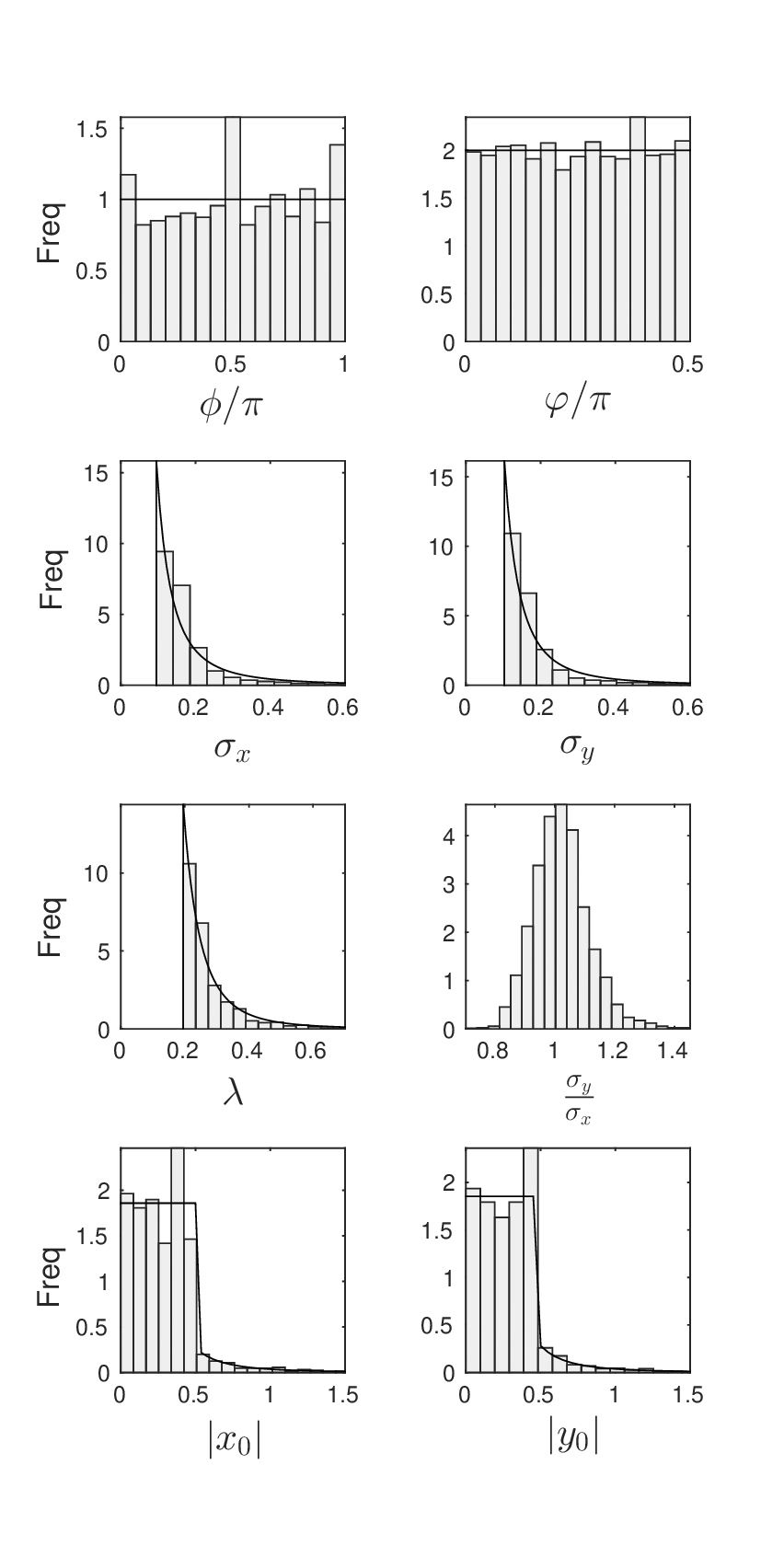}
\caption{Histograms and fitted pdfs (solid curves) of learned Gabor-function parameters. The wavelength $\lambda=2\pi/|k|$ is displayed instead of $k$ for direct comparison with other spatial parameters.}
\label{natural_histograms}
\end{figure}

Histograms and fitted probability density functions (pdfs) are shown in Fig.~\ref{natural_histograms} for the seven Gabor-function parameters. To provide the data for these histograms, ten identical models were learned independently using a different random sequence of image patches, resulting in 2560 data points per histogram (i.e., 10 models, each comprised of 256 Gabor functions). The histograms for $\phi$ and $\varphi$ extend over the full range of orientations and phases approximately uniformly. From the top-left histogram of Fig.~\ref{natural_histograms}, it can also be seen that horizontal and vertical components (i.e., $\phi=0, \pi/2$, and $\pi$) occur more frequently than others. This could either be due to the existence of more horizontal and vertical components in natural images (e.g.~trees and horizons), or an artifact of using a rectangular coordinate system \citep{vanhateren}. The parameters $x_0$ and $y_0$ (histograms of the absolute values $|x_0|$ and $|y_0|$ are shown in Fig.~\ref{natural_histograms}) extend over the whole domain of an image patch approximately uniformly. Some values also lie outside the range of an image patch as evidenced by the tails in the histograms, however, at least part of each corresponding basis function does overlap an image patch. Histograms for $\sigma_{x}$, $\sigma_{y}$, and $\lambda$ appear to be highly non-uniform with long tails. These non-uniform distributions result in the multiscale nature of the Gabor functions shown in Fig.~\ref{natural_basis}; i.e., there are many smaller Gaussian envelopes, and fewer large ones. An additional histogram, showing the Gabor-function aspect ratio $\sigma_{y}/\sigma_{x}$, is also given. This histogram is sharply peaked around one, meaning that all learned Gabor functions are close to unit aspect ratio: $\sigma_{y}/\sigma_{x}=1$. A surprising result of using a Gabor-function parameterization is that the initial mode structure of $\sigma_{y}/\sigma_{x}$ appears to be conserved by the learning rules. That is, despite $\sigma_{y}$ and $\sigma_{x}$ changing significantly during learning, their ratio remains sharply peaked around its initial value(s). This was found to be true even for multi-modal initial conditions, and may be related to the symmetry of $\sigma_{x}$ and $\sigma_{y}$ in the Gabor parameterization of $g({\bf{r}},{\bf{r}}^{\prime})$.

\begin{figure}
\center
\includegraphics[width=300pt]{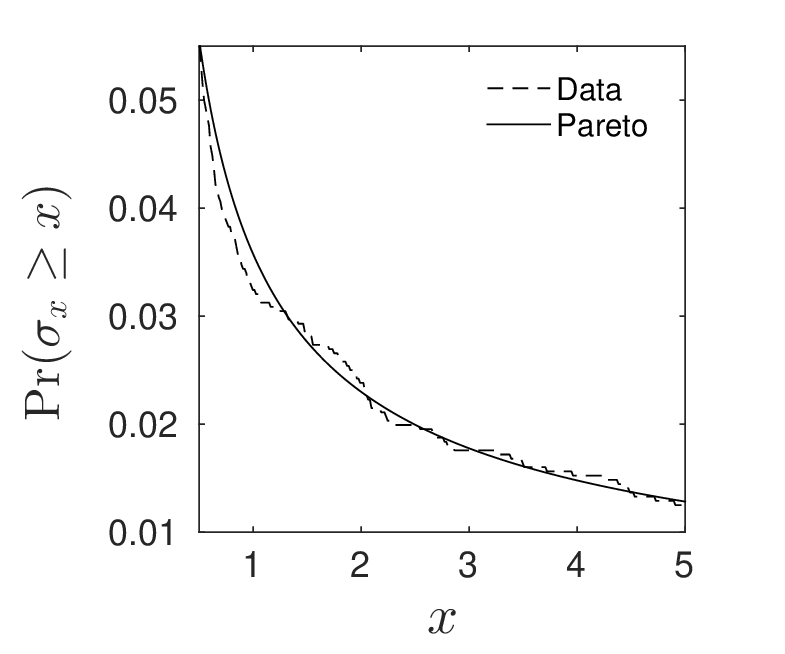}
\caption{Plot of the probability of $\sigma_x\geq x$ versus $x$ for data (dashed curve) and the Pareto tail-function (solid curve).}
\label{tail}
\end{figure}

The fitted pdfs shown in Fig.~\ref{natural_histograms} for $\phi$ and $\varphi$ correspond to normalized uniform distributions. The fitted pdfs for $\sigma_{x}$, $\sigma_{y}$, and $\lambda$ are given by the Pareto pdf:
\begin{align}
p(x|\alpha,\beta)=\left\{
\begin{array}{ll}
0 &\text{for}\ x<\beta,\\
\frac{\alpha \beta^{\alpha}}{x^{\alpha+1}} &\text{for}\ x\geq\beta.
\end{array}
\right.
\label{pareto}
\end{align}
This pdf depends on two parameters and describes a power-law distribution with a heavy tail. In Fig.~\ref{tail}, I quantify the heavy tail in $\sigma_x$ by plotting the probability that $\sigma_x\geq x$ for a range of $x$-values between $x=0.5$ and $x=5$. These values are somewhat arbitrary, but extend much deeper into the tails of the data than those shown in Fig.~\ref{natural_histograms}. The dashed curve in Fig.~\ref{tail} is given by the $\sigma_x$ data, while the solid curve is proportional to the Pareto tail-function: 
\begin{align}
Pr(X\geq x)&=\int_x^{\infty} p(y|\alpha,\beta)dy,\\ 
&= \left(\frac{\beta}{x}\right)^{\alpha},\label{tailfn}
\end{align}
with $\alpha$ and $\beta$ given by their maximum-likelihood values $\hat{\alpha}=0.637$, and $\hat{\beta}=0.501$ estimated from a dataset satisfying $\sigma_x\geq 0.5$. This was necessary because the $\sigma_x$ data not in the tail fit a different $\alpha$-value which underestimates the heaviness of the tail. The constant of proportionality ensures that $Pr(\sigma_x\geq x)$ equals the correct ratio of data points when $x=0.5$ (i.e., given by the number of data points for $\sigma_x\geq 0.5$ divided by the total number of data points). From Fig.~\ref{tail}, it is seen that the Pareto distribution fits the tail in the $\sigma_x$-data extremely well across an order of magnitude of values from 0.5 to 5. The fit is also good from 0.4 to 10 (not shown), before resolution becomes limited by the number of data points satisfying $\sigma_x> 10$. Similar observations also apply to the data for $\sigma_{y}$ and $\lambda$. The fitted pdfs for $|x_0|$ and $|y_0|$ are each a mixture of a uniform pdf and a Pareto pdf. In this case I used the EM algorithm to find maximum-likelihood estimates for the mixing coefficient $\pi$, and Pareto shape parameter $\alpha$, after fixing the value of $\beta$ to 0.5 (see Fig.~\ref{natural_histograms}).

Due to its power-law, the Pareto distribution is scale-invariant: $p(\kappa x|\alpha,\beta)=\kappa^{-(\alpha+1)} p( x|\alpha,\beta)$ for $x\geq \beta$. This can also be seen in a $\log\log$ plot of Eq.~(\ref{tailfn}) versus $x$, yielding a straight line of slope $-\alpha$. The marginal distributions for $\sigma_x$,  $\sigma_y$, and $\lambda$ are therefore scale-invariant over a wide range of values, so that the distribution of Gabor-function sizes looks the same (up to multiplication by a constant) on any length scale within this range. If simple cells have adapted to natural image statistics, then the distribution of simple-cell receptive-field sizes should also be scale-invariant over some range. That is, there is no characteristic receptive-field size selected by natural image statistics: receptive fields exist on all length scales larger than a certain minimum size.

To find parameter dependencies it is necessary to model the joint probability distribution of the Gabor-function parameters. The fitted pdfs in Fig.~\ref{natural_histograms} represent parameter marginal distributions, and can be used to transform parameter values so that they have standard-normal marginal distributions. The advantage of doing this is that dependencies in the joint distribution can be quantified using familiar multivariate Gaussian techniques. This approach also sets up the copula modelling applied in the next section. 

Transformation of Gabor-function parameters can be done by making use of the probability integral transform, followed by an inverse transform. For $\sigma_{x}$, this requires application of the cumulative distribution function (CDF) for the Pareto distribution fitted to the $\sigma_{x}$ data, followed by application of the inverse CDF for the standard normal distribution. Similarly for $\sigma_{y}$ and $\lambda$. As $\phi$ and $\varphi$ are considered to be approximately uniformly distributed, all that is required is to scale the parameter values to lie between 0 and 1, before applying the inverse CDF for the standard normal distribution. Transforming $|x_0|$ and $|y_0|$ requires the CDF for the mixture distribution fitted to the $|x_0|$ and $|y_0|$ parameters. This is straightforward to find due to the piecewise-linear and monotonic behaviour of that pdf. Following application of the mixture CDF, application of the inverse CDF for the standard normal distribution is then required. 

\begin{figure}
\center
\includegraphics[width=380pt]{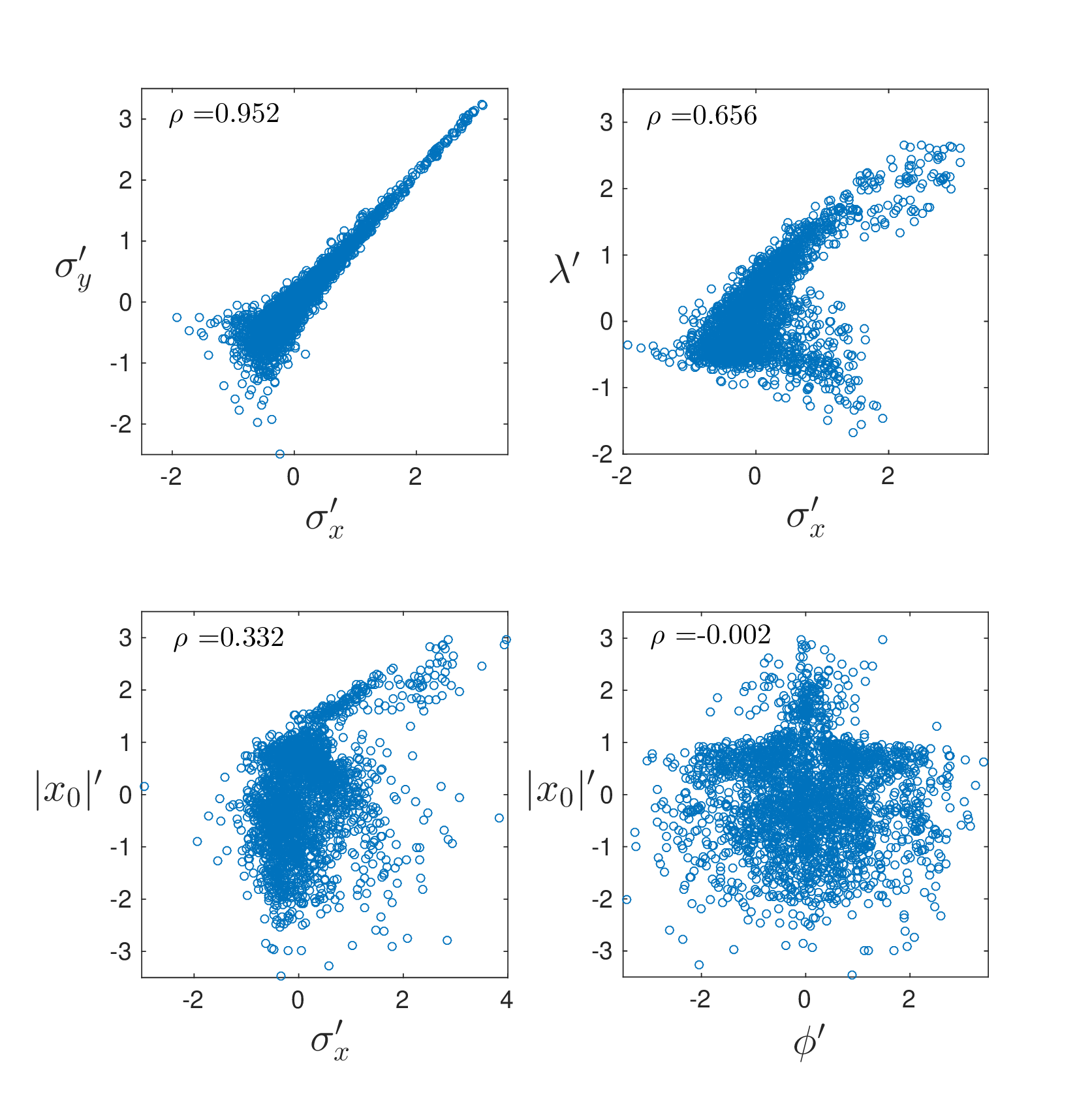}
\caption{Scatter plots and correlation coefficients for select pairs of Gabor-function parameters.}
\label{correlations}
\end{figure}

After transforming parameter marginal distributions into standard-normal form, the joint distribution of parameter values can be visualized using pairwise scatter plots. From the twenty one possible plots, four noteworthy ones are displayed in Fig.~\ref{correlations}. The largest correlation coefficient corresponds to the dependency between $\sigma_{x}^\prime$ and $\sigma_{y}^\prime$ shown in the top-left plot of Fig.~\ref{correlations} (the primed notation is used to denote the transformed parameters). This correlation shows that the aspect ratio is approximately the same for each basis function. The next-largest correlation coefficient corresponds to the dependency between $\sigma_{x}^\prime$ and $\lambda^\prime$ shown in the top-right plot of Fig.~\ref{correlations} (there is a similar dependency for $\sigma_{y}^\prime$ and $\lambda^\prime$). These correlations show that Gaussian envelopes of a given width are modulated by sinusoids of a given wavelength: larger envelopes are generally modulated by longer wavelengths. This ensures that larger basis functions do not have too many subfields, and smaller basis functions do not have too few (or zero) subfields. The third-largest correlation coefficient corresponds to the dependency between $\sigma_{x}^\prime$ and $|x_{0}|^\prime$ shown in the bottom-left plot of Fig.~\ref{correlations}. Similar dependencies also exist for the pairs $\sigma_{x}^\prime,|y_{0}|^\prime$; $\sigma_{y}^\prime,|x_{0}|^\prime$; and $\sigma_{y}^\prime,|y_{0}|^\prime$. These correlations show the width of a Gaussian envelope varies weakly with position. No other substantial correlations exist, however, \emph{all} parameter pairs exhibit some form of dependency in their scatter plots. One example is shown in the bottom-right plot of Fig.~\ref{correlations}. This plot shows that orientation varies in a complicated way with position, even though the correlation coefficient is approximately zero. 

\begin{figure}
\center
\includegraphics[width=300pt]{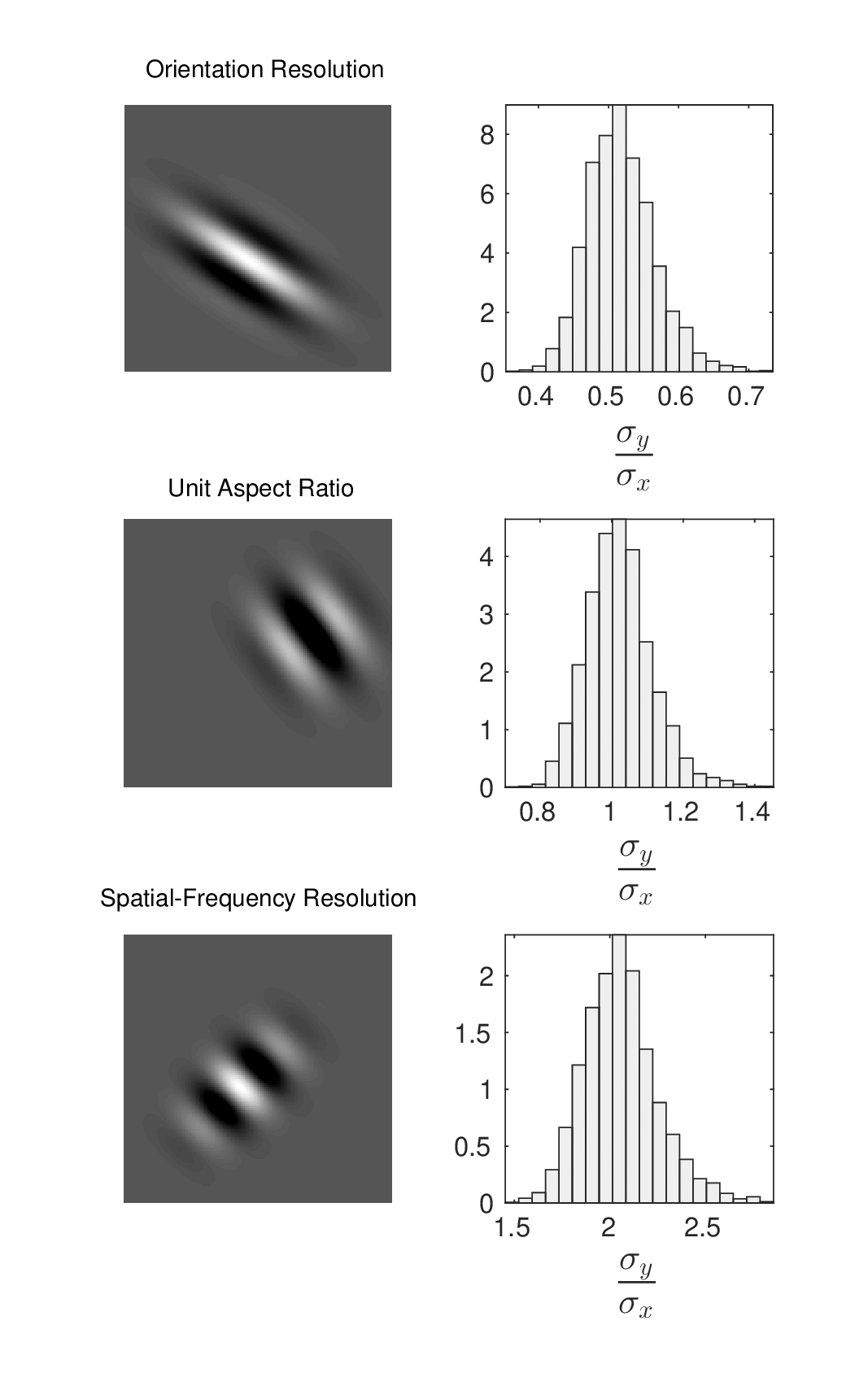}
\caption{Each row corresponds to a representative basis function (left), and a corresponding $\sigma_y/\sigma_x$ histogram (right) for a basis of 256 Gabor functions learned with specific initial parameter values. These were: $\sigma_x=0.5$, $\sigma_y=0.25$ (top row); $\sigma_x=0.5$, $\sigma_y=0.5$ (middle row); and $\sigma_x=0.25$, $\sigma_y=0.5$ (bottom row).}
\label{aspect_ratio_learned}
\end{figure}

Results presented so far correspond to the case shown in the middle row of Fig.~\ref{aspect_ratio_learned}. I have labeled this case \emph{unit aspect ratio}, as the $\sigma_y/\sigma_x$ histogram is sharply peaked around one. As previously mentioned, a surprising result of the Gabor-function parameterization is that the mode of the aspect ratio appears to be conserved by the learning rules. For this reason, applying the learning rules to the initial conditions $\sigma_x=0.5$ and $\sigma_y=0.25$ (i.e., $\sigma_y/\sigma_x=0.5$) generates the result shown in the top row of Fig.~\ref{aspect_ratio_learned}. I have labelled this case \emph{orientation resolution}, because it corresponds to sharp orientation resolution at the expense of spatial-frequency resolution when one of these Gabor functions is used as a filter. The resolution of a general 2D Gabor function in the joint 2D space and 2D spatial-frequency domains is described by the uncertainty relation \citep{daugman85},
\begin{equation}
(\Delta x\Delta y)(\Delta u\Delta v)=1/16\pi^2,\label{uncertain}
\end{equation}
where $\Delta x\Delta y$ is the effective area of a 2D Gabor function in the 2D space domain (and is proportional to $\sigma_x\sigma_y$), and $\Delta u\Delta v$ is the effective area of a 2D Gabor function in the 2D spatial-frequency domain (and is proportional to $1/\sigma_x\sigma_y$). An approximate linear relationship holds between orientation bandwidth and $\Delta u$ for small orientation bandwidths. This means that for different 2D Gabor functions with the same amount of 2D space-domain area, any gain in orientation resolution (i.e., going to smaller values of $\Delta u$) must be paid for by a loss in spatial-frequency resolution (i.e., going to larger values of $\Delta v$); and any gain in spatial-frequency resolution must be paid for by a loss in orientation resolution \citep{daugman85}. Applying the learning rules to the initial conditions $\sigma_x=0.25$ and $\sigma_y=0.5$ (i.e., $\sigma_y/\sigma_x=2$) generates the result shown in the bottom row of Fig.~\ref{aspect_ratio_learned}. I have labelled this case \emph{spatial-frequency resolution}, because it corresponds to sharp spatial-frequency resolution at the expense of orientation resolution.

\begin{figure}
\center
\includegraphics[width=350pt]{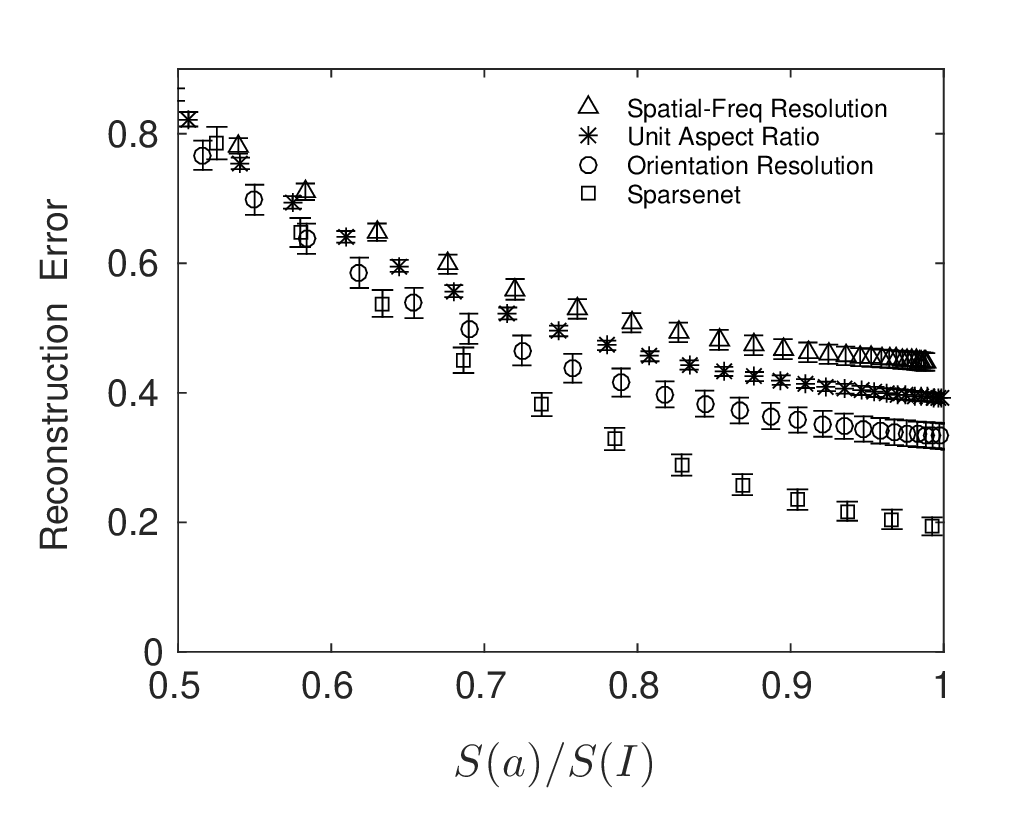} 
\caption{Average error from reconstructing 400 image patches at different sparseness levels using the different basis functions shown in Fig.~\ref{aspect_ratio_learned}. Smaller values of $S(a)/S(I)$ correspond to sparser representations (see text for details). Error bars correspond to one standard deviation.}
\label{aspect_ratio_perform}
\end{figure}

In Fig.~\ref{aspect_ratio_perform}, the average error from reconstructing 400 image patches at different sparseness levels using different types of basis function is shown. Given an input image $I({\bf{r}})$, the output activity $a({\bf{r}})$ corresponds to a sparse representation of that image. The quantity $S(a)/S(I)=\langle S(\sum_{{\bf{r}}}a({\bf{r}}))/S(\sum_{{\bf{r}}}I({\bf{r}))}\rangle$ is a measure of the sparseness of the output activity compared to the input image; where the angled brackets denote an average over a batch of 400 images. The function $S(x)=\log{(1+x^2)}$ is the same as that used in Eq.~(\ref{objective2}) to promote sparseness. When this ratio has a value less than one it indicates the output activity is sparser than the input image, which is the desired outcome of finding a sparse image code. However, using a sparse representation to reconstruct natural images from an over-complete basis leads to information loss, which is shown in Fig.~\ref{aspect_ratio_perform} as an increase in the least-squares error. The first point to note is that a basis of learned Gabor functions (top three curves) approaches, but does not quite achieve, the image-reconstruction performance of ``sparsenet" \citep{olshausen96,olshausen97} (bottom curve). Sparsenet was trained on the same image set for this comparison. The second point is that at any given sparseness level, there is only a small difference in reconstruction error separating Gabor functions for orientation resolution, spatial-frequency resolution, and with unit aspect ratio.

\begin{figure}
\center
\includegraphics[width=300pt]{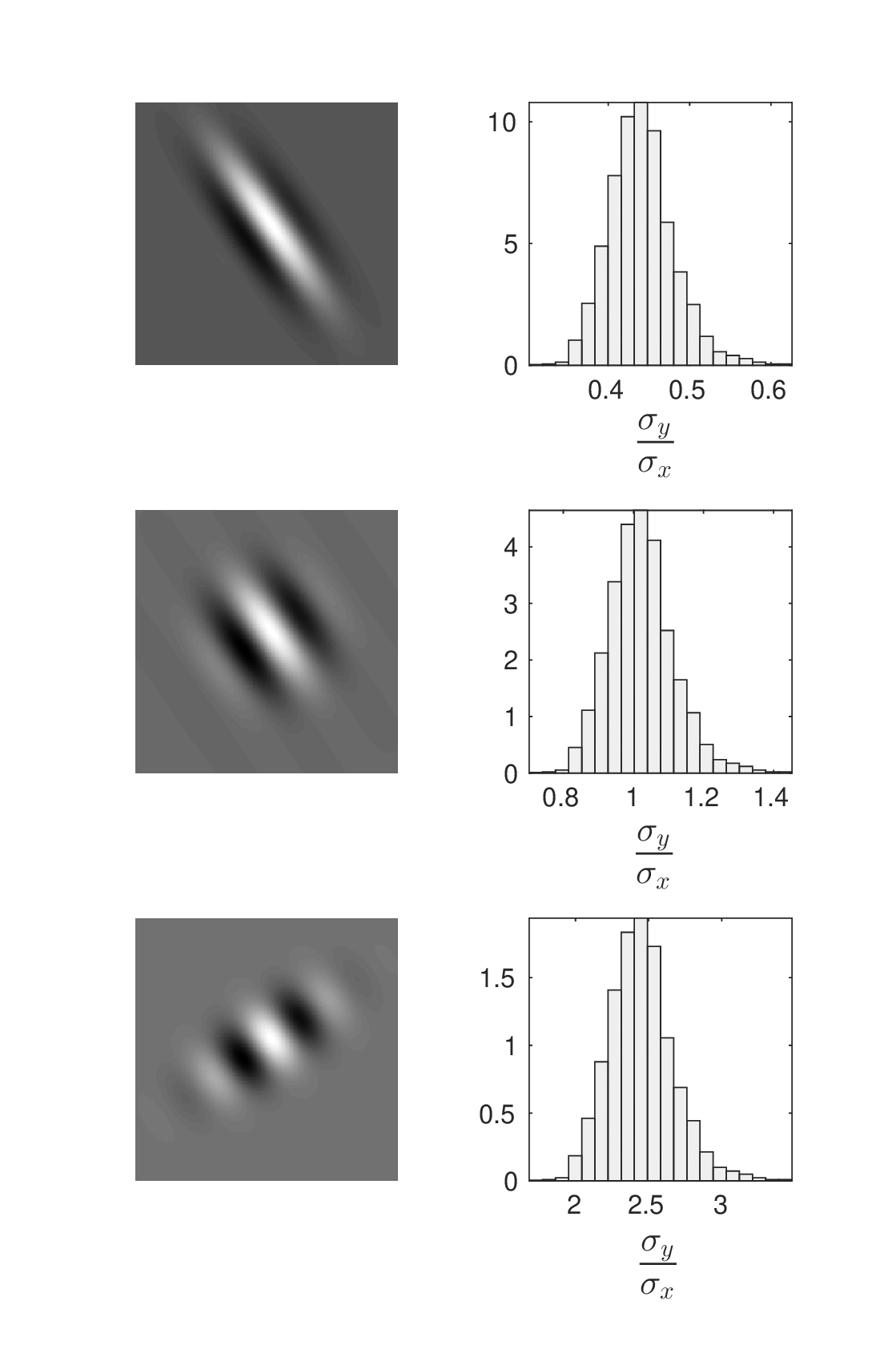}
\caption{Basis functions from using different $\beta$ values to invert the $\sigma_x^\prime$, $\sigma_y^\prime$ data from Fig.~\ref{correlations}. (Top Row) Basis functions for orientation resolution use $\beta_1=0.135$ and $\beta_2=0.063$. (Middle Row)  Basis functions for unit aspect ratio use $\beta_1=0.095$ and $\beta_2=0.103$. (Bottom Row) Basis functions for spatial-frequency resolution use $\beta_1=0.055$ and $\beta_2=0.143$.}
\label{aspect_ratio_beta}
\end{figure}

I now show that learned Gabor functions with orientation or spatial-frequency resolution can be simply related to learned Gabor functions with unit aspect ratio. In Fig.~\ref{aspect_ratio_beta}, I have constructed three different sets of basis functions from parameter data for the unit-aspect-ratio case. This was done by inverting the transformations used to get $\sigma_x^\prime$ and $\sigma_y^\prime$ (first applying the standard-normal CDF, followed by the Pareto inverse CDF), thereby returning the original parameters $\sigma_x$ and $\sigma_y$. Before inverting these transformations I changed the value of $\beta$ in the inverse CDF for $\sigma_x$ and $\sigma_y$. Using $\beta_1=0.135$ and $\beta_2=0.063$ (see Table 1 for notation) leads to basis functions for orientation resolution (top row). Using $\beta_1=0.095$ and $\beta_2=0.103$ leads to basis functions for unit aspect ratio (middle row). And using $\beta_1=0.055$ and $\beta_2=0.143$ leads to basis functions for spatial-frequency resolution (bottom row). Changing only two parameter values in the pdfs for $\sigma_x$ and $\sigma_y$ has allowed me to go continuously between orientation resolution, unit aspect ratio, and spatial-frequency resolution. Most importantly, all parameter dependencies remain unchanged, only parameter marginal distributions have been affected by these inversions. 

The results presented above are now compared with results from fitting the 2D Gabor function to simple-cell receptive-field profiles of cat \citep{daugman85,jones}, and macaque monkey \citep{vanhateren,ringach}. A key finding mentioned in all of these works is the strong correlation between the width and length of a Gabor function. Specifically, it was found that simple cells have a distribution of widths to lengths (aspect ratios) that range between 0.25 to 1 in \cite{daugman85}; 0.23 to 0.92 in \cite{jones}; 0.25 to approximately 2 in \cite{vanhateren}; and 0.29 to 5.4 in \cite{ringach}. This data suggests that in cat, simple-cell receptive-field profiles tend towards orientation resolution. In macaque, they include both orientation resolution and spatial-frequency resolution.

\begin{figure}
\center
\includegraphics[width=160pt]{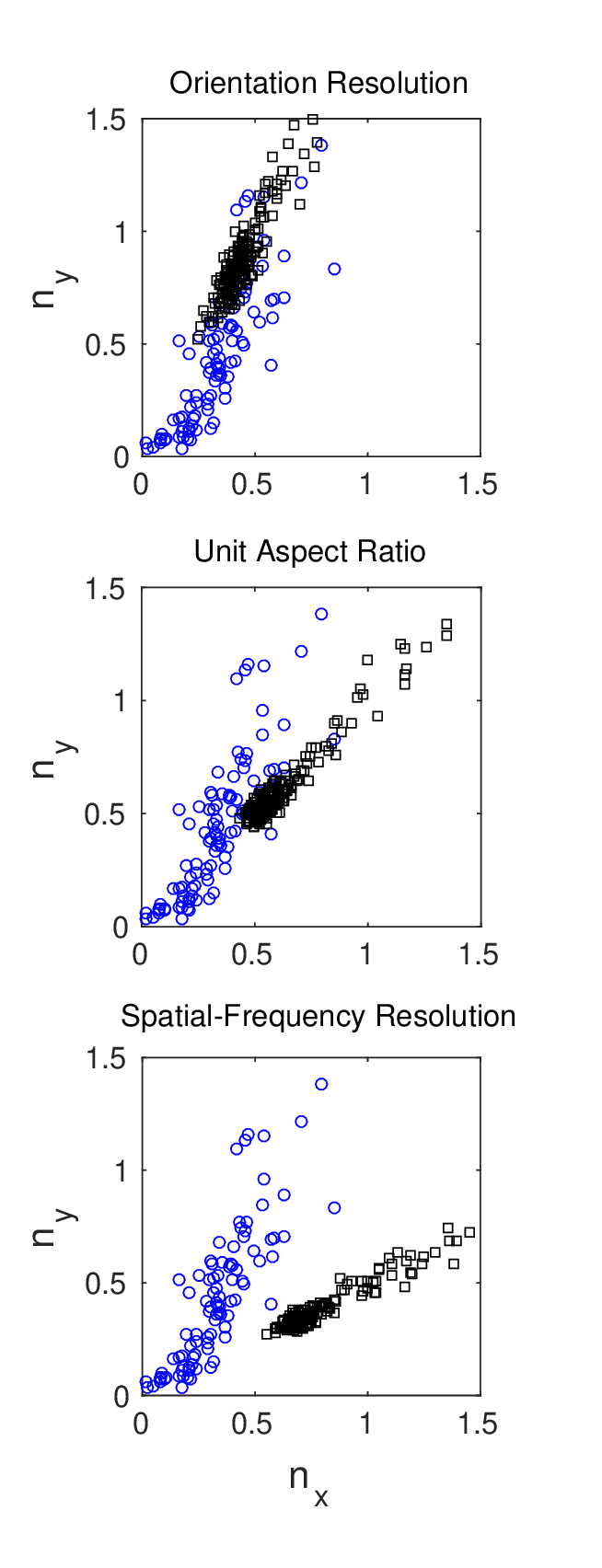}
\caption{Scatter plots of the parameters $n_{x}=\sigma_{y}/\lambda$, and $n_{y}=\sigma_{x}/\lambda$ found from fitting the Gabor function to macaque simple-cell receptive-field profiles in \cite{ringach} (circles), and from adapting the Gabor function to natural image statistics (squares) for the cases of orientation resolution (top), unit aspect ratio (middle), and spatial-frequency resolution (bottom).}
\label{macaque_data}
\end{figure}

The data from \cite{ringach} is shown in Fig.~\ref{macaque_data}, where the dimensionless quantities $n_{x}=\sigma_{y}/\lambda$, and $n_{y}=\sigma_{x}/\lambda$ have been plotted. It was necessary to swap the order of $x$ and $y$ in this comparison as the Gabor function wave-vector was aligned along the $x$-axis in \cite{ringach}. The position of data points in the $(n_{x},n_{y})$-plane reflects receptive-field profiles: cells near the origin are broadly tuned for orientation and low-pass for spatial frequency (receptive-field profiles like circular blobs), while cells away from the origin are more sharply tuned for orientation and high-pass for spatial frequency (mostly oriented receptive-field profiles with multiple subfields). I have superimposed this data onto results from adapting the Gabor function to natural image statistics. From these plots it appears the cases of orientation resolution and unit aspect ratio fit the macaque data best. It is also clear that receptive-field profiles near the origin, namely those that look like circular blobs, are not well-described by Gabor functions. Examples of sparse coding models predicting a more diverse set of receptive-field profiles are given by the works of \cite{sommer}, and \cite{olshausen13b}. 

\section{Probabilistic Generative Model}

In this section, a probabilistic model is proposed for generating 2D Gabor functions adapted to natural image statistics. The first step is to model the joint probability distribution of Gabor-function parameters from the results of Sec.~\ref{results}. The second step is to generate new values from this model, and then use Eqs.~(\ref{dic1}) and (\ref{dic2}) to find the corresponding 2D Gabor function. A set of 2D Gabor functions generated in this way can be used as a basis for sparse coding applications, or as a model of simple-cell receptive-field profiles.

To model parameter dependencies I use a Gaussian copula \citep{embrechts}. This requires inverting the parameter transformations described in Sec.~\ref{results}. Since each transformation is one-to-one and onto, it is therefore invertible. For example, the Pareto CDF can be inverted by applying the inverse CDF, and the inverse CDF for the standard normal distribution can be inverted by applying the corresponding CDF. It is important to note that some of the pre-processing steps, such as taking the absolute values of $x_0$ and $y_0$, are not invertible. These pre-processing steps will not be used in the following work. 

\begin{figure}
\center
\includegraphics[width=350pt]{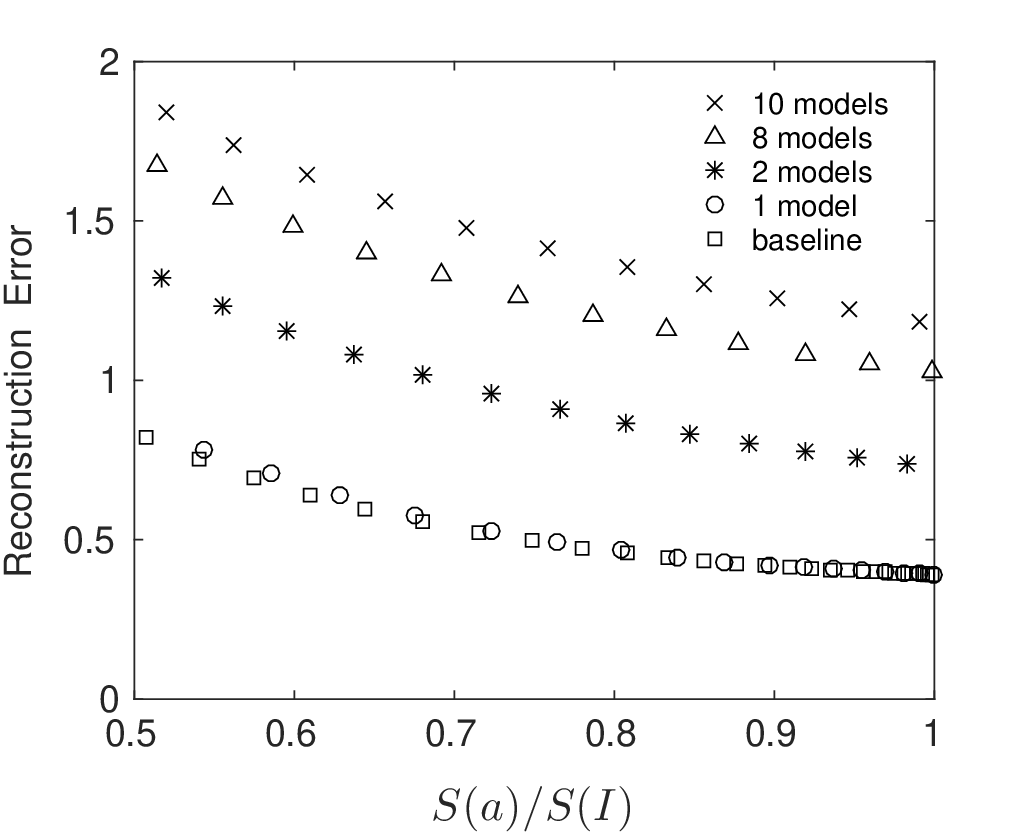}
\caption{Average error from reconstructing 400 image patches at different sparseness levels using 256 Gabor functions randomly sampled from a subset of 10 learned models (see text for details). Using all 256 Gabor functions from one model gives the plot for ``1 model". Sampling (without replacement) 256 Gabor functions from two, eight, or ten independent models gives the plots for ``2 models", ``8 models", and ``10 models", respectively.}
\label{resample}
\end{figure}

The parameter histograms in Sec.~\ref{results} represent the combined data from ten identical models that were learned independently, generating a total of 2560 data points, as previously mentioned. This was done to generate enough samples for estimating a joint probability distribution of seven parameters. For each model, the EM algorithm was applied with the sparseness-promoting term in Eq.~(\ref{objective2}), eventually leading to a set of 256 basis functions optimized for yielding sparse representations of natural images. However, combining basis functions from ten independent models now means that any arbitrary combination of 256 basis functions is unlikely to be a fixed point of the EM algorithm, and is therefore unlikely to yield an optimal sparse representation. This can be seen in Fig.~\ref{resample}, where the average error from reconstructing 400 image patches at different sparseness levels is plotted for the case of unit-aspect-ratio basis functions. The ``baseline" curve is the same as that for the unit-aspect-ratio case in Fig.~\ref{aspect_ratio_perform}, and represents the average of ten performance curves, each taken from one of the ten independent models. Now consider forming a new basis set by randomly sampling (without replacement) 256 Gabor functions from a subset of the ten models. Using all of the 256 Gabor functions from a single model leads to the data for ``1 model" plotted in Fig.~\ref{resample} that shows performance essentially as good as baseline performance. Sampling 256 Gabor functions from two models leads to the data plotted for ``2 models" in Fig.~\ref{resample}. Sampling 256 Gabor functions from eight models leads to the data plotted for ``8 models" in Fig.~\ref{resample}, and analogously for ten models. Performance is seen to degrade sharply when going from a single model to two or more models. Although combining model samples leads to reasonable estimates for parameter marginal distributions, according to Fig.~\ref{resample} this is no longer the case for parameter dependencies. For this reason I use the results of Sec.~\ref{results} to make only qualitative assumptions about the form of dependencies in the joint probability distribution.

A particularly simple model of the joint probability distribution is given using Probabilistic PCA (PPCA) \citep{tipping} to model the Gaussian copula. According to the data presented in Sec.~\ref{results}, the largest correlations exist between the three parameters $\sigma_x^\prime$, $\sigma_y^\prime$, and $\lambda^\prime$. Letting $\mathbf{x}=(\sigma_x^\prime,\sigma_y^\prime,\lambda^\prime)$, the first principal component of the data generated for $\mathbf{x}$ can be modelled using the vector $\mathbf{W}=(1,1,\rho)$. This choice includes the large symmetric correlation between $\sigma_x^\prime$ and $\sigma_y^\prime$ shown in Fig.~\ref{correlations} (which also holds for non-unit aspect ratios), as well as the correlations between $\sigma_x^\prime$ and 
$\lambda^\prime$, and $\sigma_y^\prime$ and 
$\lambda^\prime$; and which I parameterize by $\rho$. No other parameter correlations or dependencies are included in this model. Applying PPCA to generate values for $\mathbf{x}$ requires introducing a one-dimensional latent variable $z$, and the three-dimensional vectors $\boldsymbol{\mu}$ and $\boldsymbol{\epsilon}$, as 
\begin{equation}
\mathbf{x} = \mathbf{W}z + \boldsymbol{\mu} + \boldsymbol{\epsilon}.
\end{equation}\\
Assuming zero mean and zero variance outside the first principal component, this equation reduces to
\begin{equation}
\mathbf{x}=\mathbf{W}z.
\end{equation}
Random samples for $\mathbf{x}$ can be generated by sampling the latent variable from the standard normal distribution, $z\sim {\cal{N}}(0,1)$, then multiplying $z$ by $\mathbf{W}$. The corresponding probability distribution for $\mathbf{x}$ is given by a multivariate Gaussian with zero mean and covariance matrix $\mathbf{C}$:
\begin{equation}
p(\mathbf{x})={\mathcal{N}}(\mathbf{x}|0,\mathbf{C}),
\end{equation}
where,
\begin{equation}
\mathbf{C}=\mathbf{W}\mathbf{W^T}.
\end{equation}
The remaining Gabor-function parameters are modelled as independent random variables and are generated by sampling from a uniform distribution. The resulting ``PPCA-copula" generative model is summarized in Table 1. This model depends on seven free parameters controlling the shape of the marginal distributions, and the strength of correlations between $\sigma_x$, $\sigma_y$, and $\lambda$. Matlab code implementing the PPCA-copula generative model is available at \cite{loxley}.
\begin{table}
\centering
\begin{tabular}{ll}
\hline
Gabor Parameter(s)& Sample Transformation\\
\hline
$\sigma_x^\prime,\sigma_y^\prime,\lambda^\prime$&$(\sigma_x^\prime,\sigma_y^\prime,\lambda^\prime)=(1,1,\rho)z$\\
$\sigma_x$& $\sigma_x=\mathrm{PCDF}^{-1}(\mathrm{NCDF}(\sigma_x^\prime|0,1)|\alpha_1,\beta_1)$\\
$\sigma_y$& $\sigma_y=\mathrm{PCDF}^{-1}(\mathrm{NCDF}(\sigma_y^\prime|0,1)|\alpha_2,\beta_2)$\\
$\lambda$& $\lambda=\mathrm{PCDF}^{-1}(\mathrm{NCDF}(\lambda^\prime|0,1)|\alpha_3,\beta_3)$\\
$\phi$& $\phi=2\pi x$\\
$\varphi$& $\varphi=\pi x$\\
$x_0$& $x_0=-0.5+x$\\
$y_0$& $y_0=-0.5+x$\\
\hline
\end{tabular}
\vspace{10pt}
\caption{The PPCA-copula generative model. To generate one set of values for the seven Gabor-function parameters, take one sample $z\sim {\cal{N}}(0,1)$ from the standard normal distribution, and four samples $x\sim \mathrm{U}(0,1)$ from the standard uniform distribution. Then apply the parameter transformations listed in the table. Here, $\mathrm{PCDF}^{-1}(x|\alpha,\beta)=\frac{\beta}{(1-x)^{1/\alpha}}$ is the inverse CDF for the Pareto distribution, and $\mathrm{NCDF}(x|0,1)$ is the CDF for the standard normal distribution.}
\end{table}

\begin{figure}
\center
\includegraphics[width=350pt]{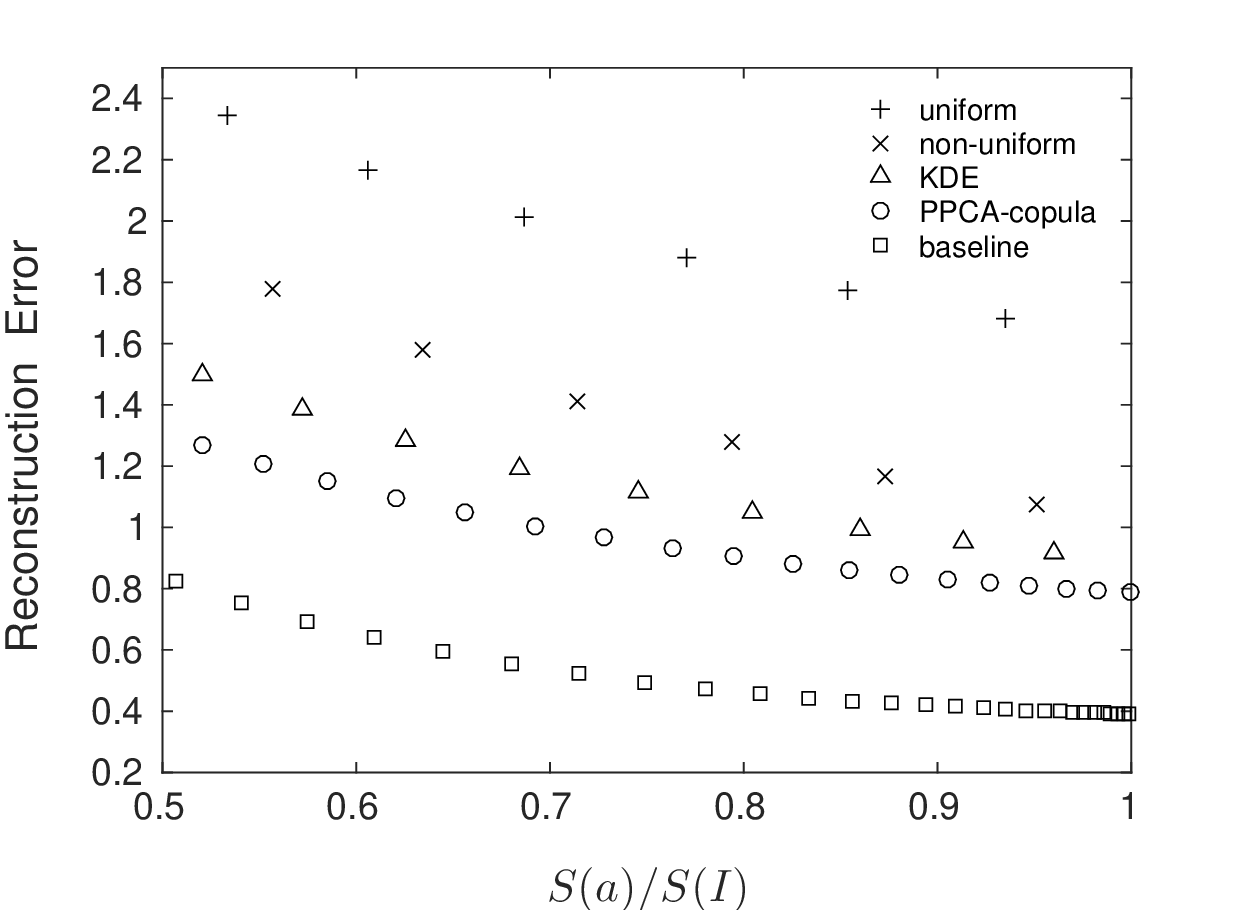}
\caption{Average error from reconstructing 400 image patches at different sparseness levels using four different generative models (see text for details).}
\label{models}
\end{figure}

The average error from reconstructing 400 image patches at different sparseness levels is shown in Fig.~\ref{models} for four different generative models. The uniform model treats parameters as statistically independent and models a uniform joint pdf by taking random samples from the uniform distribution for each of the seven Gabor-function parameters. For $\sigma_x$, $\sigma_y$, and $\lambda$, the uniform distribution sampling interval is taken as $(\mathrm{min}(i),0.4)$; where $\mathrm{min}(i)$ is the minimum data value for parameter $i$, and $0.4$ is an arbitrarily chosen realistic upper bound. The other parameter sampling schemes are as listed in Table 1. The non-uniform model also treats parameters as statistically independent, but models a non-uniform joint pdf by taking samples from the standard normal distribution for each of the seven parameters, before applying the inverse parameter transformations discussed in this section. These transformations all begin with application of the CDF for the standard normal distribution. Following that, values for $\phi$ and $\varphi$ are multiplied by $2\pi$ and $\pi$, respectively. Values for $\sigma_x$, $\sigma_y$, and $\lambda$ are transformed via the inverse CDF for the Pareto distribution, while values for $x_0$ and $y_0$ are transformed via the inverse CDF for the mixture pdf described in Sec.~\ref{results}. The KDE model includes all parameter dependencies by taking samples from a kernel density estimate \citep{ihler} of the combined ten-model data, before applying the inverse parameter transformations previously described. The PPCA-copula model is outlined in Table 1, and in  Fig.~\ref{models} uses the maximum-likelihood estimates from unit-aspect-ratio data: $ (\hat{\alpha}_1,\hat{\beta}_1,\hat{\alpha}_2,\hat{\beta}_2,\hat{\alpha}_3,\hat{\beta}_3)= (1.51,0.095,1.66,0.103,2.81,0.195)$. The value of the correlation parameter was chosen as $\rho=1.25$. 

\begin{figure}
\center
\includegraphics[width=250pt,bb=89 51 600 510, clip=true]{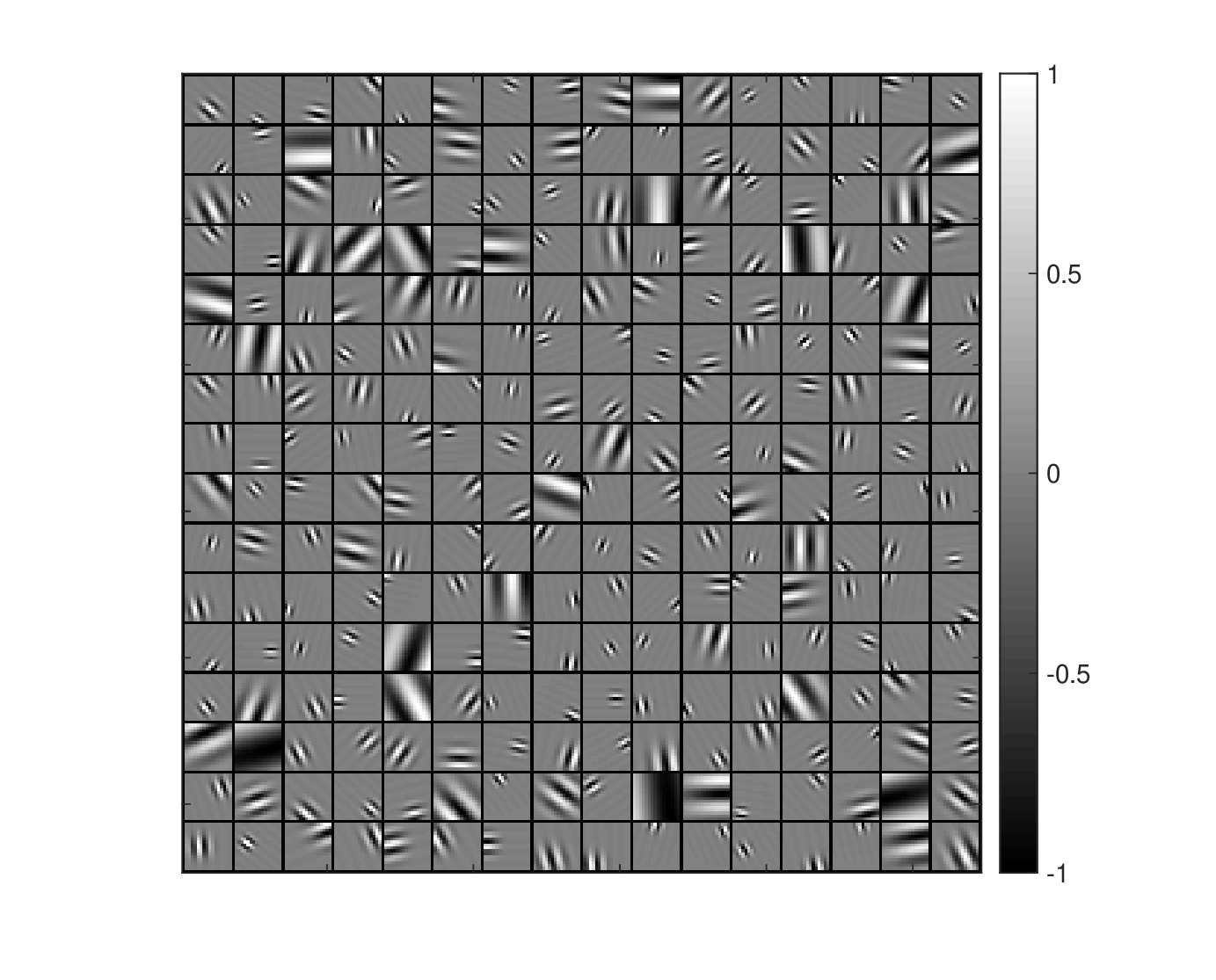}
\caption{A basis of 256 Gabor functions generated from the PPCA-copula model with $ (\alpha_1,\beta_1,\alpha_2,\beta_2,\alpha_3,\beta_3,\rho)= (1.51,0.095,1.66,0.103,2.81,0.195,1.25)$. }
\label{gen_basis}
\end{figure}

\begin{figure}
\center
\includegraphics[width=250pt,bb=90 50 600 507, clip=true]{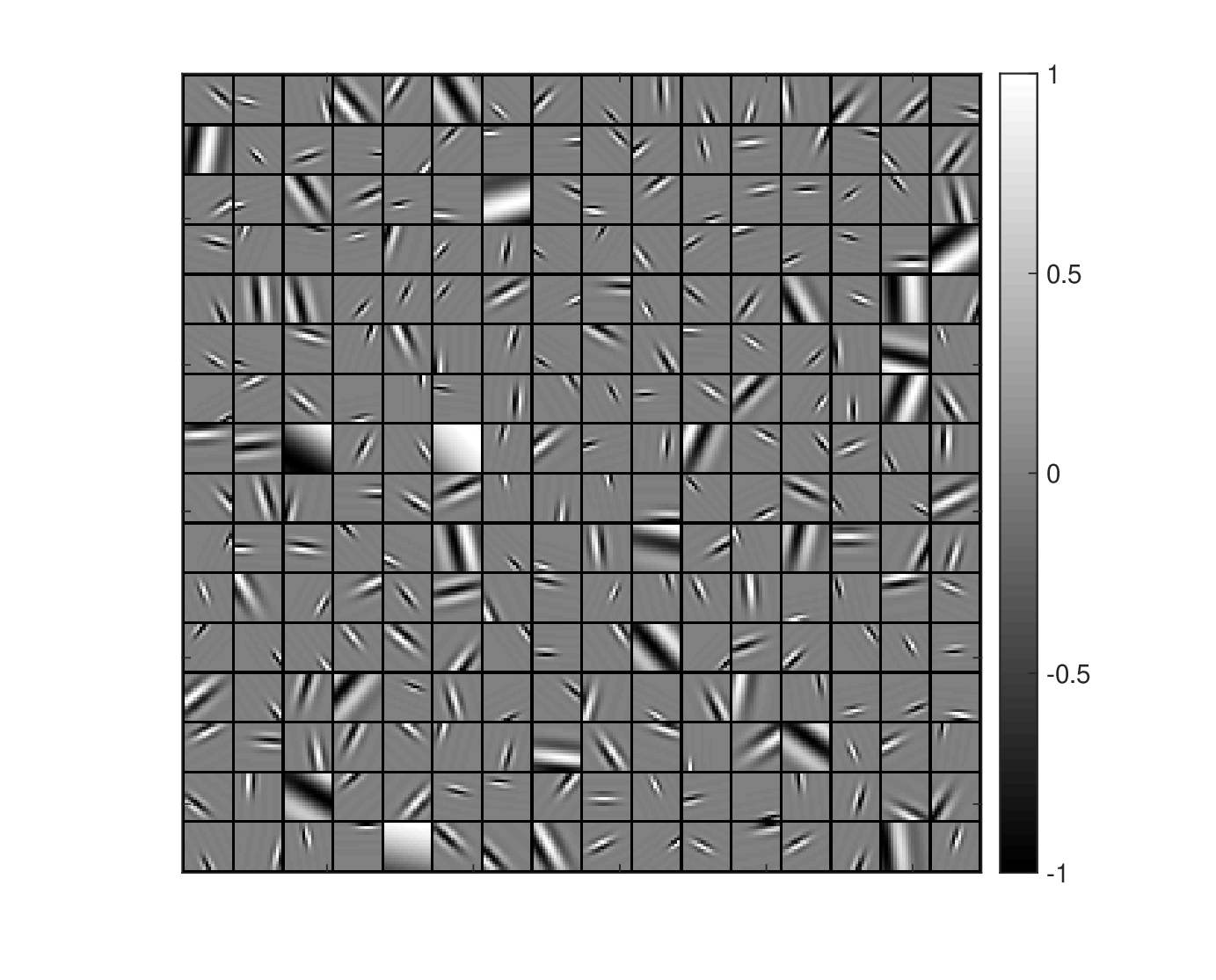}
\caption{A basis of 256 Gabor functions generated from the PPCA-copula model with the same parameter values as in Fig.~\ref{gen_basis}, except that now $\beta_1=0.130$, and $\beta_2=0.068$.}
\label{gen_basis2}
\end{figure}

In Fig.~\ref{models}, a clear model ranking can be seen. The best-performing model is given by the PPCA-copula model, followed by the KDE model, the non-uniform model, and the uniform model. A significant gap still remains between the best-performing model, and the baseline curve given by the learned Gabor-function parameters. The single largest increase in performance results from modelling the Pareto pdf when going from the uniform to non-uniform models. A smaller performance increase results from including some of the parameter dependencies in the form of the KDE or PPCA-copula models. The KDE model does not perform particularly well because parameter dependencies were modelled using the combined ten-model data. The PPCA-copula model does slightly better because only the marginal distributions were quantitatively fit to this data. Performance of the PPCA-copula model could most likely be improved by directly optimizing its parameters for image-reconstruction performance. In Figs.~\ref{gen_basis} and \ref{gen_basis2}, two different basis sets of Gabor functions have been generated using the PPCA-copula model. The basis generated in Fig.~\ref{gen_basis} corresponds to the learned basis in Fig.~\ref{natural_basis}. The basis generated in Fig.~\ref{gen_basis2} uses the same parameter values as in Fig.~\ref{gen_basis}, except $\beta_1$ and $\beta_2$ were chosen for orientation resolution rather than for unit aspect ratio. 

\section{Conclusion} 

The 2D Gabor function has previously been used to describe the linear response of simple cells to a visual stimulus, and as an image-code primitive in sparse representations of natural images. In this work, I showed the 2D Gabor function can be adapted to natural image statistics by learning a joint probability distribution for the Gabor-function parameters. The resulting over-complete basis of Gabor functions forms a sparse representation of natural images. The learned joint probability distribution was found to be characterized by heavy tails, a few strong correlations, and many weak dependencies. Three Gabor-function parameters representing the size and spatial frequency of the 2D Gabor function were shown to be responsible for much of this behaviour. These parameters are characterized by non-uniform marginal distributions with heavy tails, most likely due to scale-invariance in natural images, and all three parameters are strongly correlated: resulting in a basis of multiscale Gabor functions with similar aspect ratios, and size-dependent spatial frequencies. I quantified the heavy tails using the Pareto tail function, and concluded that a power-law distribution is necessary for describing this data. An important consequence is that the distribution of receptive-field sizes is scale-invariant over a wide range of values. This means there is no characteristic size for a receptive field that is selected by natural image statistics. Receptive fields can exist on all length scales greater than a certain minimum size; although in the case of biological receptive fields, physiological constraints would determine a maximum possible size.

The Gabor-function aspect ratio was found to be approximately conserved by the learning rules, and is therefore not well-determined by natural image statistics. This allowed for three distinct solutions: a basis of Gabor functions with sharp
orientation resolution at the expense of spatial-frequency resolution; a basis of Gabor
functions with sharp spatial-frequency resolution at the expense of orientation resolution; or a basis with unit aspect ratio. Solutions with arbitrary mixtures of all three cases are also possible. Two parameters controlling the shape of the marginal distributions in a probabilistic generative model were shown to fully account for the three distinct solutions. These findings are partly supported by the work of \cite{eichhorn}, where it was found that ``non-oriented" filters perform nearly as well as optimal orientation-selective filters at redundancy reduction within a linear model framework.

The learned joint probability distribution was approximated to yield probabilistic generative models of receptive-field profiles and basis functions for sparse coding. The best-performing model with lowest reconstruction error at a given sparseness level was found to be a Gaussian copula with Pareto marginal probability density functions. This model captures some of the strong correlations and heavy-tail characteristics of the learned joint distribution, and in the current setting suggests that learning a sparse code for natural images primarily results in a collection of multiscale basis functions with a power-law distribution of sizes. To reconstruct a natural image, a sparse coding model is therefore making use of a linear superposition of ``statistically-independent" basis functions (according to the factorial assumption usually applied in sparse coding models) that have a power-law distribution of sizes. These are the two key properties of natural images suggested by \cite{ruderman}, and it is now seen that both properties are present in sparse coding models.

Including learned dependencies between Gabor-function parameters was shown to improve image reconstruction performance, but only up to a point. The best-performing probabilistic generative model included a qualitative form of the first principal component of three key Gabor-function parameters. Going beyond this did not result in any further improvement. The reason was that Gabor-function parameters were learned in batches of 256 values, and combining different batches to increase the number of data points available for fitting parameter distributions ended up degrading the parameter dependencies. Instead of combining independent batches of parameter values, a better model would be found by learning one large batch of values. This could be done by learning a larger basis from larger image patches, but would be computationally expensive. Another improvement to this work would be to introduce a mixed parameterized basis such as a mixture of Gabor functions and difference-of-Gaussians functions. Learning a highly over-complete basis, as in \cite{sommer} and \cite{olshausen13b}, should then lead to a sparser code and a more diverse set of receptive-field profiles.

\subsection*{Acknowledgments}
I thank Luis Bettencourt, Garrett Kenyon, Reid Porter, and the staff and students of the Redwood Center for Theoretical Neuroscience at UC Berkeley for useful feedback on this work. This work was partly supported by the Center for Nonlinear Studies at Los Alamos National Laboratory, and partly supported by the U.S. Department of Energy through the LANL/LDRD Program projects 20090006DR, and 20130013DR.

\section*{Appendix}

\subsection{Learning Rules}
To derive a set of learning rules for adapting the 2D Gabor-function parameters to the statistics of natural images I follow an approach due to \cite{olshausen97}, and \cite{lewicki99}. In this approach, the probability of generating a particular image $I$ is assumed to be given by a continuous latent variable model of the form:
\begin{equation}
P(I|{\boldsymbol{\theta}})=\int da P(I|{\boldsymbol{\theta}},a)P(a),\label{imagemod}
\end{equation}
where $a$ are a set of unobserved (latent) variables, and where, for the parameterization used here, ${\boldsymbol{\theta}}({\bf{r}})=(\phi({\bf{r}}),\varphi({\bf{r}}),\sigma_{x}({\bf{r}}),\sigma_{y}({\bf{r}}),k({\bf{r}}),x_0({\bf{r}}),y_0({\bf{r}}))$ is a vector of the seven Gabor parameters for each basis function. In the case of Gaussian noise $N({\bf{r}})$ with variance $\sigma^{\prime}$: $P(I|{\boldsymbol{\theta}},a)\propto\prod_{{\bf{r}}}\exp{(-N({\bf{r}})^2/2{\sigma^{\prime}}^2)}$, with $N({\bf{r}})=I({\bf{r}})-\sum_{{\bf{r}}^{\prime}}g({\bf{r}},{\bf{r}}^{\prime}) a({\bf{r}}^{\prime})$ from Eq.~(\ref{gen}). The marginal distribution for $a$ is assumed to be sparse and to factor: $P(a)\propto\prod_{{\bf{r}}}\exp{(-\beta S(a({\bf{r}})))}$; where $S(x)=\log{(1+x^2)}$ for the Cauchy distribution is assumed here. Other common choices for $P(a)$ are the logistic distribution, and the Laplacian distribution.

Estimating parameters in a latent variable model can be done efficiently using the EM algorithm \citep{dempster}. The E step begins by inferring the latent variables $a$, given ${\boldsymbol{\theta}}$ and $I$. Using Bayes' rule, $P(a|I,{\boldsymbol{\theta}})$ can be written as
\begin{equation}
P(a|I,{\boldsymbol{\theta}})\propto P(I|{\boldsymbol{\theta}},a)P(a).\label{posterior}
\end{equation}
However, the expectation over $P(a|I,{\boldsymbol{\theta}})$ cannot be evaluated analytically, so approximate inference must be used. One approach often used in sparse coding is to assume the Maximum Posterior (MAP) estimate for $a$. Upon defining $E=-\log{[P(I|{\boldsymbol{\theta}},a)P(a)]}$, and using Eq.~(\ref{posterior}), this can be written as
\begin{eqnarray}
\hat{a}&=&\underset{a}{\operatorname{arg\ max}}\ P(a|I,{\boldsymbol{\theta}}),\label{map}\\
&=&\underset{a}{\operatorname{arg\ min}}\ E,\label{mapminE}
\end{eqnarray} 
where
\begin{equation}
E=\sum_{\bf{r}}\left\{\frac{1}{2}\left[I({\bf{r}})-\sum_{{\bf{r}}^{\prime}}g({\bf{r}},{\bf{r}}^{\prime})a({\bf{r}}^{\prime})\right]^{2}+\nu S(a({\bf{r}}))\right\},\label{objective}
\end{equation}
with $\nu={\sigma^{\prime}}^2\beta$. In this equation all terms independent of $a$ and $g$ (the density normalizations do not depend on $a$ or $g$) have been neglected, and $E$ has been re-scaled by ${\sigma^{\prime}}^2$. Finding the MAP value for $a$ therefore reduces to simultaneously minimizing the least-squares error and sparseness terms in Eq.~(\ref{objective}). In general this is a non-convex optimization, since the second derivative of $\log{(1+a^2)}$ with respect to $a$ becomes negative outside $a\in(-1,1)$. Therefore, any minimum of $E$ is not guaranteed to be a global minimum \citep{boyd}. This can result in multi-modal posterior distributions, as shown pictorially in \cite{seeger}. However, in practice it seems the correct minimum is found efficiently using conjugate gradient descent, and experiments using convex forms for $S(a)$ show results that are qualitatively similar to those for $S(a)=\log{(1+a^2)}$ \citep{olshausen96}.

The M step involves maximizing $\langle\log{P(I|{\boldsymbol{\theta}})}\rangle$ with respect to the Gabor parameters ${\boldsymbol{\theta}}$. This average log-likelihood is given by the log of the likelihood function in Eq.~(\ref{imagemod}), averaged over a batch of images. Maximizing the average log-likelihood is equivalent to minimizing the Kullback-Leibler divergence between the distribution of images in nature, and the distribution of images generated from the image model \citep{olshausen97}. This can be implemented using gradient ascent:
\begin{eqnarray}
\Delta\theta_{i}({\bf{r}})&=&\eta_{i}\frac{\partial }{\partial \theta_{i}({\bf{r}})}\langle \log{P(I|{\boldsymbol{\theta}})}\rangle,\\\nonumber \\
&=&\eta_{i} \left\langle \frac{1}{P(I|{\boldsymbol{\theta}})} \frac{\partial }{\partial \theta_{i}({\bf{r}})}\int da P(I|{\boldsymbol{\theta}},a)P(a)\right\rangle,\label{parteqn}
\end{eqnarray}
where $\eta_{i}$ are the Gabor parameter learning rates, and Eq.~(\ref{imagemod}) has been used for $P(I|{\boldsymbol{\theta}})$. It is now convenient to use $P(I|{\boldsymbol{\theta}},a)P(a)=\exp{(-E)}$ from the definition of $E$, allowing Eq.~(\ref{parteqn}) to be written as
\begin{eqnarray}
\Delta\theta_{i}({\bf{r}})&=&\eta_{i} \left\langle \frac{1}{P(I|{\boldsymbol{\theta}})} \int da P(I|{\boldsymbol{\theta}},a)P(a)\left (-\frac{\partial E}{\partial \theta_{i}({\bf{r}})}\right )\right\rangle,\\\nonumber \\
&=&-\eta_{i} \left\langle \int da P(a|{\boldsymbol{\theta}},I)\frac{\partial E}{\partial \theta_{i}({\bf{r}})}\right\rangle,\\\nonumber \\
&=&-\eta_{i} \left\langle\left\langle \frac{\partial E}{\partial \theta_{i}({\bf{r}})}\right\rangle_{P(a|{\boldsymbol{\theta}},I)}\right\rangle.\label{firstLR}
\end{eqnarray}
The gradient of $E$ is given by
\begin{equation}
\frac{\partial E}{\partial \theta_{i}({\bf{r}})}=-a({\bf{r}})\sum_{\bf{r}^{\prime}} r({\bf{r}}^{\prime})\frac{\partial g({\bf{r}}^{\prime},{\bf{r}})}{\partial \theta_{i}({\bf{r}})},
\end{equation}
where the residual error $r({\bf{r}})$ is defined as
\begin{equation}
r({\bf{r}})=I({\bf{r}})-\sum_{{\bf{r}}^{\prime}}g({\bf{r}},{\bf{r}}^{\prime})a({\bf{r}}^{\prime})\label{error}.
\end{equation}
Using these two expressions in Eq.~(\ref{firstLR}) leads to
\begin{equation}
\Delta\theta_{i}({\bf{r}})=\eta_{i} \sum_{{\bf{r}}^{\prime}} \frac{\partial g^{T}({\bf{r}},{\bf{r}^{\prime}})}{\partial\theta_{i}({\bf{r}})} \left\langle \left\langle a({\bf{r}})r({\bf{r}}^{\prime})\right\rangle_{P(a|{\boldsymbol{\theta}},I)}\right\rangle.\label{twoexp}
\end{equation}
This is similar to the learning rule of \cite{olshausen97} except for the partial derivative term, which allows each Gabor parameter to be updated independently. The partial derivatives are provided in the next section.

Updating each Gabor parameter therefore requires the calculation of two expectations. The inner expectation in Eq.~(\ref{twoexp}) is with respect to the posterior distribution $P(a|{\boldsymbol{\theta}},I)$ given by Eq.~(\ref{posterior}) and comprises the E step. The outer expectation is an average over a batch of images. Adjusting each Gabor parameter according to Eq.~(\ref{twoexp}) is the M step. The EM algorithm consists of alternating between the E step and the M step until convergence is reached \cite[e.g., see][]{bishop}.

If the noise level is zero and the basis is complete, the E step can be avoided and the ICA learning rule follows. Then Eq.~(\ref{gen}) can be inverted to give $a({\bf{r}})=\sum_{{\bf{r}}^{\prime}}g({\bf{r}},{\bf{r}}^{\prime})^{-1} I({\bf{r}}^{\prime})$, and the distribution $P(I|{\boldsymbol{\theta}},a)$ in Eq.~(\ref{imagemod}) becomes a delta-function over $a$. Performing the integral over $a$ in Eq.~(\ref{imagemod}) then yields $P(I|{\boldsymbol{\theta}})=P(g^{-1}I)$. Maximizing $\langle\log{P(I|{\boldsymbol{\theta}})}\rangle$ with respect to $g^{-1}$ forms the basis of the FastICA algorithm \citep{hyvarinenbook}.

In the presence of Gaussian noise and an over-complete basis, the E step is usually performed either by sampling from $P(a|{\boldsymbol{\theta}},I)$, or by using its MAP estimate from Eq.~(\ref{mapminE}). For the case of the MAP estimate, the learning rule in Eq.~(\ref{twoexp}) becomes
\begin{equation}
\Delta\theta_{i}({\bf{r}})=\eta_{i} \sum_{{\bf{r}}^{\prime}} \frac{\partial g^{T}({\bf{r}},{\bf{r}^{\prime}})}{\partial\theta_{i}({\bf{r}})} \left\langle \hat{a}({\bf{r}})\hat{r}({\bf{r}}^{\prime})\right\rangle,\label{maplearn}
\end{equation}
where $\hat{r}$ is the residual error from Eq.~(\ref{error}) with $\hat{a}$ instead of $a$, and $\hat{a}$ is the MAP estimate given by Eqs.~(\ref{mapminE}) and (\ref{objective}). The drawback of using MAP in the E step is encountering a trivial solution given as follows. The area occupied by a basis function increases with increase in $\sigma_{x}({\bf{r}^{\prime}})$ or $\sigma_{y}({\bf{r}^{\prime}})$. This results in an increase in the value of $\sum_{{\bf{r}}} |g({\bf{r}},{\bf{r}^{\prime}})|^2$ for a particular ${\bf{r}^{\prime}}$. Now both terms in Eq.~(\ref{objective}) can be minimized by a small value of $a({\bf{r}^{\prime}})$, and a large value of $\sum_{{\bf{r}}} |g({\bf{r}},{\bf{r}^{\prime}})|^2$; such that $\sum_{{\bf{r}}} |g({\bf{r}},{\bf{r}^{\prime}})a({\bf{r}^{\prime}})|^2\approx \sum_{{\bf{r}}}|I({\bf{r}})|^2$ for a particular value of ${\bf{r}^{\prime}}$. One way to avoid a trivial solution where $\sigma_{x}({\bf{r}^{\prime}})$ and $\sigma_{y}({\bf{r}^{\prime}})$ simply move to larger values is to approximate the envelope of each Gabor function as an ellipse: $\tilde{x}^2/\sigma_{x}({\bf{r}})^2+\tilde{y}^2/\sigma_{y}({\bf{r}})^2=1$, and make use of the formula for the area of an ellipse,
\begin{equation*}
A_{\mathrm{ellipse}}({\bf{r}})=\pi\sigma_{x}({\bf{r}})\sigma_{y}({\bf{r}}).
\end{equation*}
Updates of $\sigma_{x}({\bf{r}})$ and $\sigma_{y}({\bf{r}})$ can then be appropriately constrained by modifying a rule used in \cite{olshausen97} that makes use of the variance of $\hat{a}({\bf{r}})$, as:
\begin{equation}
A_{\mathrm{ellipse}}({\bf{r}})^{\mathrm{new}}=A_{\mathrm{ellipse}}({\bf{r}})^{\mathrm{old}}\left[\frac{\langle \hat{a}({\bf{r}})^2\rangle}{\sigma_{\mathrm{goal}}^{2}}\right]^{\alpha},\label{modOF}
\end{equation}
followed by
\begin{equation}
\sigma_{x}({\bf{r}})^{\mathrm{new}}= \frac{A_{\mathrm{ellipse}}({\bf{r}})^{\mathrm{new}}}{\pi\sigma_{y}({\bf{r}})^{\mathrm{old}}},
\end{equation}
if the parameter grouping $(\sigma_x,x_0)$ is being updated, or
\begin{equation}
\sigma_{y}({\bf{r}})^{\mathrm{new}}= \frac{A_{\mathrm{ellipse}}({\bf{r}})^{\mathrm{new}}}{\pi\sigma_{x}({\bf{r}})^{\mathrm{old}}},\label{ellipse}
\end{equation}
if the parameter grouping $(\sigma_y,y_0)$ is being updated (there is no further update for the grouping $(\phi,\varphi,k)$). When the variance of $\hat{a}({\bf{r}})$ over an image batch falls below $\sigma_{\mathrm{goal}}^{2}$, $\sigma_{x}({\bf{r}})$ and $\sigma_{y}({\bf{r}})$ then decrease according to Eqs.~(\ref{modOF})--(\ref{ellipse}). This update causes the variance of $\hat{a}({\bf{r}})$ to increase, with the learning rule finding a fixed point when $\sigma_{\mathrm{goal}}^{2}$ is reached. Therefore, this heuristic prevents the variance of $\hat{a}({\bf{r}})$ from becoming too small, and $\sigma_{x}({\bf{r}})$ and $\sigma_{y}({\bf{r}})$ from becoming too large. The set of learning rules applied in Sec.~3 consist of Eq.~(\ref{mapminE}) (the E step) and Eqs.~(\ref{maplearn})--(\ref{ellipse}) (the M step).

\subsection{Partial Derivatives for Learning Rules}

The partial derivatives required in Eqs.~(\ref{twoexp}) and (\ref{maplearn}) are given here:
\begin{eqnarray}
\frac{\partial g^{T}({\bf{r}},{\bf{r}^{\prime}})}{\partial\phi({\bf{r}})}&=&g^{T}({\bf{r}},{\bf{r}^{\prime}})\left(\frac{1}{\sigma_{x}({\bf{r}})^2}-\frac{1}{\sigma_{y}({\bf{r}})^2}\right)\tilde{x}\tilde{y}-h^{T}({\bf{r}},{\bf{r}^{\prime}})k({\bf{r}})\tilde{x},\\\nonumber\\
\frac{\partial g^{T}({\bf{r}},{\bf{r}^{\prime}})}{\partial\sigma_{x}({\bf{r}})}&=&g^{T}({\bf{r}},{\bf{r}^{\prime}})\frac{\tilde{x}^2}{\sigma_{x}({\bf{r}})^3},\\\nonumber\\
\frac{\partial g^{T}({\bf{r}},{\bf{r}^{\prime}})}{\partial\sigma_{y}({\bf{r}})}&=&g^{T}({\bf{r}},{\bf{r}^{\prime}})\frac{\tilde{y}^2}{\sigma_{y}({\bf{r}})^3},\\\nonumber\\
\frac{\partial g^{T}({\bf{r}},{\bf{r}^{\prime}})}{\partial k({\bf{r}})}&=&-h^{T}({\bf{r}},{\bf{r}^{\prime}})\tilde{y},\\\nonumber\\
\frac{\partial g^{T}({\bf{r}},{\bf{r}^{\prime}})}{\partial\varphi({\bf{r}})}&=&-h^{T}({\bf{r}},{\bf{r}^{\prime}}),\\\nonumber\\
\frac{\partial g^{T}({\bf{r}},{\bf{r}^{\prime}})}{\partial x_0({\bf{r}})}&=&g^{T}({\bf{r}},{\bf{r}^{\prime}})\left(\cos{\phi{(\bf{r})}}\frac{\tilde{x}}{\sigma_{x}({\bf{r}})^2}+\sin{\phi{({\bf{r}})}}\frac{\tilde{y}}{\sigma_{y}({\bf{r}})^2}\right)+h^{T}({\bf{r}},{\bf{r}^{\prime}})k({\bf{r}})\sin{\phi{({\bf{r}})}},\\\nonumber\\
\frac{\partial g^{T}({\bf{r}},{\bf{r}^{\prime}})}{\partial y_0({\bf{r}})}&=&g^{T}({\bf{r}},{\bf{r}^{\prime}})\left(\cos{\phi{(\bf{r})}}\frac{\tilde{y}}{\sigma_{y}({\bf{r}})^2}-\sin{\phi{({\bf{r}})}}\frac{\tilde{x}}{\sigma_{x}({\bf{r}})^2}\right)+h^{T}({\bf{r}},{\bf{r}^{\prime}})k({\bf{r}})\cos{\phi{({\bf{r}})}},
\end{eqnarray}
where $g^{T}({\bf{r}},{\bf{r}^{\prime}})=g({\bf{r}^{\prime}},{\bf{r}})$ is the transpose of $g({\bf{r}},{\bf{r}^{\prime}})$, and $h^{T}({\bf{r}},{\bf{r}^{\prime}})=h({\bf{r}^{\prime}},{\bf{r}})$ is the transpose of
\begin{equation}
h({\bf{r}},{\bf{r}}^{\prime})=A\exp{\left[-\frac{1}{2}\left(\frac{\tilde{x}^2}{\sigma_{x}({\bf{r}}^{\prime})^2}+\frac{\tilde{y}^2}{\sigma_{y}({\bf{r}}^{\prime})^2}\right)\right]}\sin{\left[k({\bf{r}}^{\prime})\tilde{y}+\varphi({\bf{r}}^{\prime})\right]},
\end{equation}
and $(\tilde{x},\tilde{y})$ are defined as in Eq.~(\ref{dic2}).

\end{document}